\documentclass[pdflatex,sn-apa]{sn-jnl}

\usepackage{graphicx}
\renewcommand{\orcidlogo}{%
  \includegraphics[width=10pt]{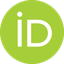}%
}
\renewcommand{\orcid}[1]{%
  \href{https://orcid.org/#1}{\orcidlogo}%
}
\usepackage{multirow}
\usepackage{amsmath,amssymb,amsfonts}
\usepackage{amsthm}
\usepackage{xcolor}
\usepackage{textcomp}
\usepackage{manyfoot}
\usepackage{booktabs}

\begin{document}

\title[Virtual metrology for phototransistor gain]{Forward and Inverse Virtual
Metrology for Phototransistor Gain: A Hierarchical, Uncertainty-Aware Approach
for Small Production Datasets}

\author[1,2]{\fnm{Mahshid} \sur{Amirabgir}\orcid{0009-0006-4849-2624}}\email{mahshid.amirabgir@students.uniroma2.eu}
\author[1]{\fnm{Lorenza} \sur{Ferrario}\orcid{0000-0003-2175-4306}}\email{ferrario@fbk.eu}
\author[3]{\fnm{Paolo} \sur{Conci}\orcid{0000-0003-4011-7695}}\email{paolo.conci@microfabsolutions.com}
\author[4]{\fnm{Mahdieh} \sur{Amirabgir}\orcid{0009-0005-0615-1058}}\email{mahdieh.amirabgir@mail.polimi.it}
\author*[2]{\fnm{Giancarlo} \sur{Orengo}\orcid{0000-0003-1044-2999}}\email{orengo@ing.uniroma2.it}

\affil[1]{\orgdiv{Micro Nano Facility (MNF), Center for Sensors and Devices},
  \orgname{Fondazione Bruno Kessler (FBK)},
  \orgaddress{\street{Via Sommarive 18}, \city{Povo (Trento)},
  \postcode{38123}, \country{Italy}}}
\affil[2]{\orgdiv{Department of Electronic Engineering},
  \orgname{University of Rome Tor Vergata},
  \orgaddress{\street{Via del Politecnico 1}, \city{Rome},
  \postcode{00133}, \country{Italy}}}
  \affil[3]{\orgname{Microfab Solutions},
  \orgaddress{\street{Via Sommarive 18}, \city{Povo (Trento)},
  \postcode{38123}, \country{Italy}}}
\affil[4]{\orgdiv{Department of Civil and Environmental Engineering (DICA)},
  \orgname{Politecnico di Milano},
  \orgaddress{\street{Piazza Leonardo da Vinci 32}, \city{Milan},
  \postcode{20133}, \country{Italy}}}

\abstract{The customization, optimization and stabilization of the process flow
of a silicon bipolar phototransistor commits months of cleanroom time before a
finished device can be measured, so a model that predicts device gain from
process parameters before a run has value out of proportion to its accuracy. We
study this problem on a real fabrication history, thirteen to fourteen
process runs of a single device: a small-sample, hierarchically structured
setting unlike the large-corpus regime of conventional virtual metrology.
Decomposing the variance of device gain, we find that roughly half of it lies
between process runs rather than within them, so recipe-only prediction is
bounded by construction. Building on these findings we provide a forward gain
predictor with a relative, uncertainty-aware signal, an inverse search that
returns recipes for a target gain, and, as the foundation for all of it, a
multi-level data-quality assessment tailored to the nested physical entities of
fabrication (batch, wafer, die) with an explicit cross-level linkage score. The
normalized dataset and analysis code are released for full reproducibility.}


\maketitle

\section{Introduction}\label{sec:intro}

In the strategic panorama of semiconductors, supported by the EU Commission with
investments and research frameworks such as the Chips Act and the recently
announced Chips Act 2, a significant role is played by research institutions
actively contributing to the ``lab to fab'' path. Moving from pure research and
development to the market requires several steps, often corresponding to a
gradual increase in Technology Readiness Level (TRL), where highly innovative
laboratory devices are optimized to become part of application systems based on
stabilized, repeatable fabrication process flows. In addition to improving the
quality of a manufacturing process, specific customization may be necessary in
order to adapt the process itself to a specific application.

Two examples explain this concept: silicon photomultipliers (SiPM) and silicon
bipolar phototransistors. SiPM can be adapted to different light wavelengths by
working on the components known as ``optical windows'', allowing SiPM for the
ultraviolet (UV) or near-infrared--ultraviolet (NIR--UV) ranges, or, combined
with germanium, for the infrared (IR) spectrum.

Silicon bipolar phototransistors, the device this paper focuses on, can be
addressed to industrial applications such as decoders, which require a specified
collector--emitter breakdown voltage ($BV_{CE}$), or to applications requiring
controlled, high response speed.

Simulation, design and fabrication of a silicon phototransistor customized to a
specific application \citep{dallabetta1998} commits months of cleanroom time and
irreversible material before a single finished device can be measured. A process
engineer choosing an emitter implant dose or an oxidation step is therefore
making a decision whose consequence, the current gain of the resulting device,
will not be known until a full fabrication cycle has completed. This asymmetry,
between the speed of a decision and the cost of learning whether it was right,
is what makes predictive modelling of fabrication outcomes valuable: a model
that forecasts gain from process parameters before a run, even imperfectly, can
convert blind iteration into informed choices and remove clearly infeasible
recipes before they consume fab time.

Predicting product quality from process parameters rather than measuring it
directly is the established domain of virtual metrology \citep{vmreview2024}.
That literature, however, has developed largely in high-volume processes where
tens of thousands of wafers are available, and where natural groupings in the
data (chambers, tools, products) are commonly handled by training a separate
model for each group, an approach that limits generalization across groups
\citep{breidung2025}. The setting of this paper is the opposite on both counts.
The data come from a real but limited fabrication history, on the order of
fourteen process runs of a single device. This situation is typical in pilot
lines, where new processes are repeated until controlled repeatability is
reached and are then transferred to production, or when small productions are
required to cover niche markets. The problem is therefore not how to exploit a
large corpus but how to model honestly when batch-level data are scarce. And
rather than fit a separate model per run, we treat the grouping structure
explicitly, which turns out to be not merely a modelling convenience but the
route to the paper's central scientific result.

That result is a finding about where gain variation lives. Decomposing the
variance of device gain shows that roughly half of it sits \emph{between}
process runs rather than within them: the run, not only the recipe, governs the
outcome. A direct consequence is that any model using only recipe parameters,
which are fixed within a run, is bounded from above in what it can predict, and
we report that bound honestly rather than obscuring it. A second, subtler
finding concerns the effect of a wafer's position within its run. During
fabrication, wafers are loaded into horizontal furnaces to activate the
formation of a thin layer of silicon dioxide. Typically 25 wafers are loaded on
a quartz carrier which enters the furnace. A wafer's index within its run has an
effect on gain that is substantial within individual runs, but its direction is
not shared across runs, so a single pooled model averages it toward a small,
unrepresentative value and understates how much position matters within any
given run. We do not claim this index corresponds to a physical position in the
thermal chamber, as the load order is not recorded (Section~\ref{sec:missing});
we treat it as within-run sequence order throughout. Recovering this requires modelling each run's behaviour explicitly,
and we confirm it through several independent statistical views that agree.

These findings rest on data whose trustworthiness must itself be established,
and fabrication data poses a quality problem that flat, single-table assessment
does not address: it is organized in strict physical strata, batch, wafer and
die, and the failures that matter most are failures of the correspondence
\emph{between} levels. We therefore develop a multi-level data-quality
assessment tailored to these physical entities, with an explicit cross-level
linkage score, as the foundation the modelling stands on.

\begin{figure}[htbp]
\centering
\includegraphics[width=\textwidth]{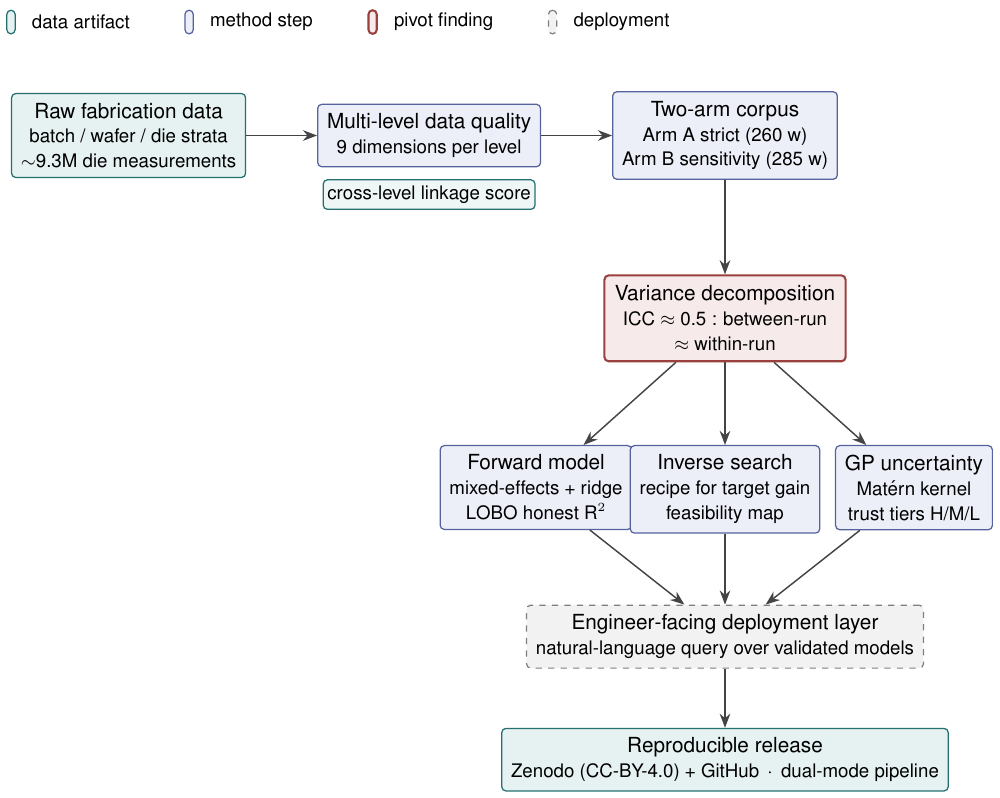}
\caption{End-to-end methodology. Raw multi-level fabrication data is assessed by
the multi-level data-quality framework, including a cross-level linkage score,
yielding a two-arm corpus. Variance decomposition (intraclass correlation
approximately 0.5) is the pivot finding that motivates the modelling choices: a
forward mixed-effects model, an inverse recipe search and a Gaussian-process
uncertainty layer. The validated models are exposed through a deployment layer,
and the normalized dataset and dual-mode pipeline are released for
reproducibility. Boxes are keyed by outline style and fill as shown in the
legend: data artifact, method step, pivot finding and deployment}
\label{fig:workflow}
\end{figure}

The main contributions of this paper are three, and
Fig.~\ref{fig:workflow} summarizes how they fit together as an end-to-end
methodology.

First, a \emph{prediction and recipe-design capability} for device gain: a
forward model that predicts gain from process parameters with
leave-one-batch-out honest accuracy and a relative uncertainty signal, an
inverse search that returns the recipes capable of reaching a target gain, and a
Gaussian-process uncertainty layer that tells an engineer how far to trust each
prediction. We are explicit that the cross-run predictive accuracy is modest, a
direct and expected consequence of the variance structure above, and frame the
models as an honest screening aid rather than a replacement for fabrication.

Second, a \emph{scientific finding} about the process: that between-run variance
dominates recipe effects (the intraclass correlation, in the sense of
\citet{nakagawa2017}, is near one-half), and that the within-run position effect
is real but sign-inconsistent across runs, a structure invisible to any single
pooled model and recovered only by modelling the run explicitly.

Third, a \emph{multi-level data-quality methodology} that extends per-level
quality scoring to the nested physical entities of fabrication data and
introduces an explicit cross-level linkage score, with each level's dimensions
chosen, and each exclusion justified, rather than applied uniformly.

A reproducible public release of the normalized wafer-level dataset is deposited
on Zenodo under a CC-BY-4.0 licence \citep{amirabgir2026data} and the analysis
code on GitHub
(\url{https://github.com/mahshid-amirabgir/phototransistor-virtual-metrology}),
with a dual-mode pipeline that produces identical results on the released
normalized data and the internal raw data, so every result here is reproducible
from the public artifacts.

Beyond these contributions, we operationalize the prediction capability through
an engineer-facing natural-language assistant
(Section~\ref{sec:assistant}). A local large language model interprets an
engineer's question and calls the validated forward, inverse and uncertainty
models to answer it, but is constrained so that it never generates a gain value
or a recipe itself; every number it returns originates from the models and is
shown with its provenance. This is a deployment layer rather than a new
modelling result: its purpose is to make the validated models usable by a
non-specialist without exposing the modelling machinery, and to run entirely on
a local machine so that no proprietary recipe data leaves the institute.

The remainder of the paper is organized as follows.
Section~\ref{sec:related} reviews related work in phototransistor fabrication,
data-driven process modelling, hierarchical models, Gaussian-process uncertainty
and data quality. Section~\ref{sec:data} describes the device, the process and
the dataset. Section~\ref{sec:dq} presents the multi-level data-quality
assessment. Section~\ref{sec:models} presents the models and findings, the
variance hierarchy and position effect, then the forward, inverse and
uncertainty models. Section~\ref{sec:assistant} presents the engineer-facing
natural-language assistant. Section~\ref{sec:discussion} discusses implications
and generalization.

\section{Related work}\label{sec:related}

\subsection{Phototransistor fabrication and prior FBK work}\label{sec:rw-device}

The silicon bipolar phototransistor is a mature optoelectronic device: a bipolar
junction transistor with an enlarged base--collector junction that collects light
and amplifies the resulting photocurrent through transistor action, yielding
responsivities well above those of comparable photodiodes. The device studied
here descends directly from the optical-encoder phototransistor developed at the
institute (then IRST, now Fondazione Bruno Kessler) and optimized by
\citet{dallabetta1998}, who used coupled process and device simulation to define
a fabrication sequence achieving current gains of several hundred and
responsivities above 0.4~A/W, and established that the emitter implant is the
dominant lever on gain. That work is the foundation for the present device and
process.

The phototransistor remains an active research target, though recent work has
moved largely toward novel materials and structures rather than the conventional
implanted-silicon device. High-performance variants have been demonstrated using
indium phosphide nanopillars grown on silicon for optical interconnects
\citep{ko2016}, thin InGaAs films on silicon waveguides reaching responsivities
of order $10^{6}$~A/W for photonic-circuit power monitoring \citep{ochiai2022},
reconfigurable silicon-nanowire phototransistors operating in both n- and p-type
polarities for broadband UV--visible--NIR optical sensing on CMOS-compatible
photonic platforms \citep{sinanowire2026}, and GeSn/Ge heterojunction bipolar
phototransistors on silicon for optical-network applications \citep{gesn2026}.
These advances target device performance and integration.

Our contribution is orthogonal to this device-physics line. The conventional
silicon bipolar process is established and deployed, and its limiting practical
problem is no longer device design but the cost and variability of customization
and of fabrication. We therefore bring a data-driven modelling approach to the
established process, predicting and inverting gain from process parameters
across many real fabricated batches, rather than proposing a new device.

\subsection{Data-driven process modelling in semiconductor fabrication}\label{sec:rw-vm}

Predicting product quality from process and equipment data, rather than
measuring it directly, is the established paradigm of virtual metrology (VM) and
advanced process control. A recent systematic review \citep{vmreview2024} defines
VM as the use of machine-learning conjecture models that take tool-state, recipe
and sampled metrology variables as inputs to predict fabrication quality
targets, and surveys its application across the major fab process steps; the
motivation throughout is that physical metrology is costly and slow, so a
reliable prediction reduces the measurement burden and tightens control. This is
precisely the motivation of the present work, and our forward gain model is a VM
model in this sense.

Two features of the mainstream VM literature distinguish it from the setting
here. First, it is typically a large-corpus enterprise: VM is well developed in
high-volume, high-throughput processes such as chemical-mechanical planarization
(CMP), plasma etch, and chemical vapor deposition (CVD), where large datasets
and dense sensor traces are available to train flexible models
\citep{dailey2024}. Second, where the data contain natural groupings
(chambers, products), it is common to train separate models per group, which a
recent study on CMP non-uniformity prediction notes often limits cross-group
generalizability \citep{breidung2025}; the broader use of machine learning across
CMP is surveyed comprehensively by \citet{winkler2025}. That CMP study also
illustrates the arc this paper follows, repurposing a forward predictive model
toward inverse recipe optimization. A common practical convention in this
literature, which we also adopt, is to normalize features and withhold absolute
recipe values for confidentiality; while many studies simply discard samples with
missing values, an unpublished CVD virtual-metrology study instead benchmarks
imputation as preprocessing and reports that it improves accuracy (Xie \&
Stearrett, preprint arXiv:2107.05071).

Our contribution sits deliberately apart from large-corpus VM on two axes. The
data are small where modelling decisions are made, on the order of fourteen
fabrication runs, so the question is not how to exploit a large corpus but how to
model honestly when the corpus is small. And rather than train separate models
per run to handle the grouping, we model the grouping \emph{explicitly} through
hierarchical random effects, which both uses the data more efficiently and turns
the between-group structure from a nuisance into the central scientific finding
(Section~\ref{sec:models}).

\subsection{Mixed-effects and hierarchical models for process data}\label{sec:rw-mixed}

When data are organized in nested groups, a model that ignores the grouping
treats correlated observations as independent and misstates both effects and
their uncertainty. Mixed-effects (multilevel, hierarchical) models address this
by adding random effects for the grouping factors alongside the fixed effects of
interest, partitioning variance into between-group and within-group components
rather than pooling them. This structure fits semiconductor fabrication data
naturally, where batch, wafer and die form intrinsic nested levels. A recent
review of regression and predictive modelling in wafer manufacturing notes that
such multi-level data structures and correlated measurements are common in
high-volume manufacturing, and that hierarchical and mixed-model approaches are
used for yield and wafer-map prediction \citep{waferreview2025}. Directly
relevant to the present setting, a hierarchical variability model for
silicon-photonic circuits separates systematic from random process variation
across the wafer, die and device levels \citep{xing2023photonics}, consistent
with the nested lot--wafer--die grouping our analysis relies on, where lots
contain at most twenty-five wafers and often fewer.

The statistics that make such models interpretable are well established. The
intraclass correlation coefficient (ICC), also called the variance partition
coefficient, quantifies the share of total variance attributable to a grouping
factor; for a random-intercept model it is the between-group variance divided by
the total \citep{nakagawa2017}. Marginal and conditional $R^{2}$, generalized to
generalized linear mixed models by \citet{nakagawa2013} and extended to
random-slope models by \citet{johnson2014}, separate the variance explained by
the fixed effects alone (marginal) from that explained by fixed and random
effects together (conditional); their difference reflects the variance
attributable to the random effects. \citet{nakagawa2017} revisit and expand these
measures to further distributional families. When no fixed effects are fitted,
the ICC equals the gap between conditional and marginal $R^{2}$, the same
quantity viewed two ways.

These tools are central to this paper. Section~\ref{sec:models} begins by
partitioning gain variance with the ICC, finds that roughly half is between-run,
and uses exactly the marginal-versus-conditional contrast to show that the
within-run position effect is real but run-specific. Because that effect is a
random \emph{slope}, not merely a random intercept, the random-slope extensions
above are what make the reported $R^{2}$ values appropriate. What such models do
not by themselves provide is a calibrated, point-by-point predictive uncertainty
for a new recipe; for that we turn to Gaussian processes.

\subsection{Gaussian-process uncertainty quantification in manufacturing}\label{sec:rw-gp}

A mixed-effects model partitions variance and yields confidence intervals on its
parameters, but it does not by itself provide a point-by-point predictive
uncertainty that widens as a query moves away from the observed data.
Gaussian-process regression (GPR) does. A Gaussian process (GP) is a
non-parametric Bayesian model that places a distribution over functions and
returns, for every input, a predictive mean and variance \citep{rasmussen2006};
the variance grows in regions far from the training data, so the model reports
not only a prediction but where it lacks confidence. Two properties make GPR well
suited to the present setting. It performs well with small datasets, since it
conditions directly on the available observations rather than estimating many
parameters, and its uncertainty is intrinsic to the probabilistic framework
rather than bolted on; an unpublished tutorial emphasizes precisely that GPs
accommodate the uncertainty arising from insufficient data and make explicit
where the model is unsure (Li \& Wang, preprint arXiv:2502.03090). Its behaviour
is governed chiefly by
the covariance kernel, with the squared-exponential and Mat\'ern families the
common choices, so the kernel is a modelling decision rather than a default.

GPR has been used in semiconductor virtual metrology to attach predictive
uncertainty to a quality target. It has been applied to predict material-removal
rate in chemical-mechanical planarization with quantified uncertainty
\citep{cai2020} and, closest to the setting here, in a recent online
Gaussian-process method that operates from minimal initial metrology data and
uses the predictive uncertainty itself to decide where measurement is needed
\citep{han2025}. In both, the predictive variance is a deliberate output rather
than a by-product; in the latter it is acted on directly, the logic this paper
adopts in turning GP variance into engineer-facing trust tiers.

In Section~\ref{sec:models} we use a Gaussian process over the recipe features,
with a Mat\'ern kernel, as the uncertainty layer over the forward model: its
predictive standard deviation is translated into engineer-facing trust tiers, and
its divergence between interpolation and leave-one-run-out prediction is read
directly as the signature of the between-run variance established in
Section~\ref{sec:variance}. How the trustworthiness of the underlying data is
itself assessed, the precondition for any of this modelling, is the subject of
the data-quality literature.

\subsection{Data-quality methodologies for fabrication data}\label{sec:rw-dq}

Data quality has been formalized through dimension-based frameworks that
decompose fitness for use into measurable axes. \citet{wangstrong1996} derived
data-quality dimensions empirically from data consumers and grouped them into
intrinsic, contextual, representational and accessibility categories,
establishing the dimensional vocabulary the field still reuses; recent surveys of
quality metrics and tooling for machine learning organize dimensions along the
same lines and note that few available tools are tailored to machine-learning
tasks \citep{zhou2024}. Several recent frameworks address this fragmentation: a
materials data quality governance framework arranges nine dimensions across a
machine-learning lifecycle, pairing data-driven methods with domain knowledge to
detect and correct issues \citep{liu2025a}; an extension of the ISO/IEC~25012
standard consolidates the literature's many terms into a hierarchical model and
adds governance, usefulness, quantity and semantics dimensions
\citep{miller2024}; and others score up to ten dimensions and combine them
through a fuzzy comprehensive evaluation model with weights derived by the
analytic hierarchy process (AHP), demonstrated on tabular datasets such as
KDDCup99 \citep{zhang2024}.

Two threads are directly relevant here. On small, high-dimensional data, where
the feature-space dimension is large relative to the sample size,
\citet{liu2023} argue the feature-to-sample ratio should itself be a governance
target, cautioning that generative augmentation may produce samples that are not
physically valid for a given material system and advocating domain-knowledge
constraints to keep generated data within reasonable value ranges; we adopt the
ratio-based view in scoring quantity and, in our setting, forgo augmentation
altogether. On anomaly detection, \citet{liu2025b} encode domain knowledge as
symbolic rules spanning individual descriptor values, inter-descriptor
correlations and inter-sample similarity; we adapt the
individual-descriptor-value rule for the accuracy dimension. Broader reviews map
data quality against the machine-learning lifecycle \citep{priestley2023} and
catalog manufacturing-adapted metrics \citep{peixoto2025}, while a recent
production framework embeds real-time scoring and drift detection in
continuous-stream pipelines \citep{bayram2026}, the latter premised on a data
stream rather than the fixed historical corpus assessed here.

A smaller line of work measures quality at more than one structural level, and it
is the closest prior work. Multi-level tools compute dimensions at the schema
level of an information system \citep{ehrlinger2018schema}, or across nested
aggregation levels (attribute, concept, data source, system) with per-level
scoring under user-adjustable per-dimension weights \citep{illescas2021}, and a
recent survey of measurement tools reports that cross-level aggregation is
supported by almost no existing tool and never above the table level
\citep{ehrlinger2022}. In every case, however, the levels are strata of a
database or information system. No prior work targets the nested \emph{physical}
entities of semiconductor fabrication (batch, wafer, die), nor, to our
knowledge, formalizes a quantitative cross-level linkage score between adjacent
levels. Those two gaps are the contributions of the multi-level data-quality
assessment developed in Section~\ref{sec:dq}.

\section{Data}\label{sec:data}

This section describes the device and its fabrication, the process parameters
used as model inputs, how device gain is defined, and the hierarchical structure
of the dataset on which the models of Section~\ref{sec:models} are built.

\subsection{Phototransistor fabrication overview}\label{sec:device-fab}

\begin{figure}[htbp]
\centering
\begin{minipage}{0.32\textwidth}\centering
\includegraphics[width=\linewidth]{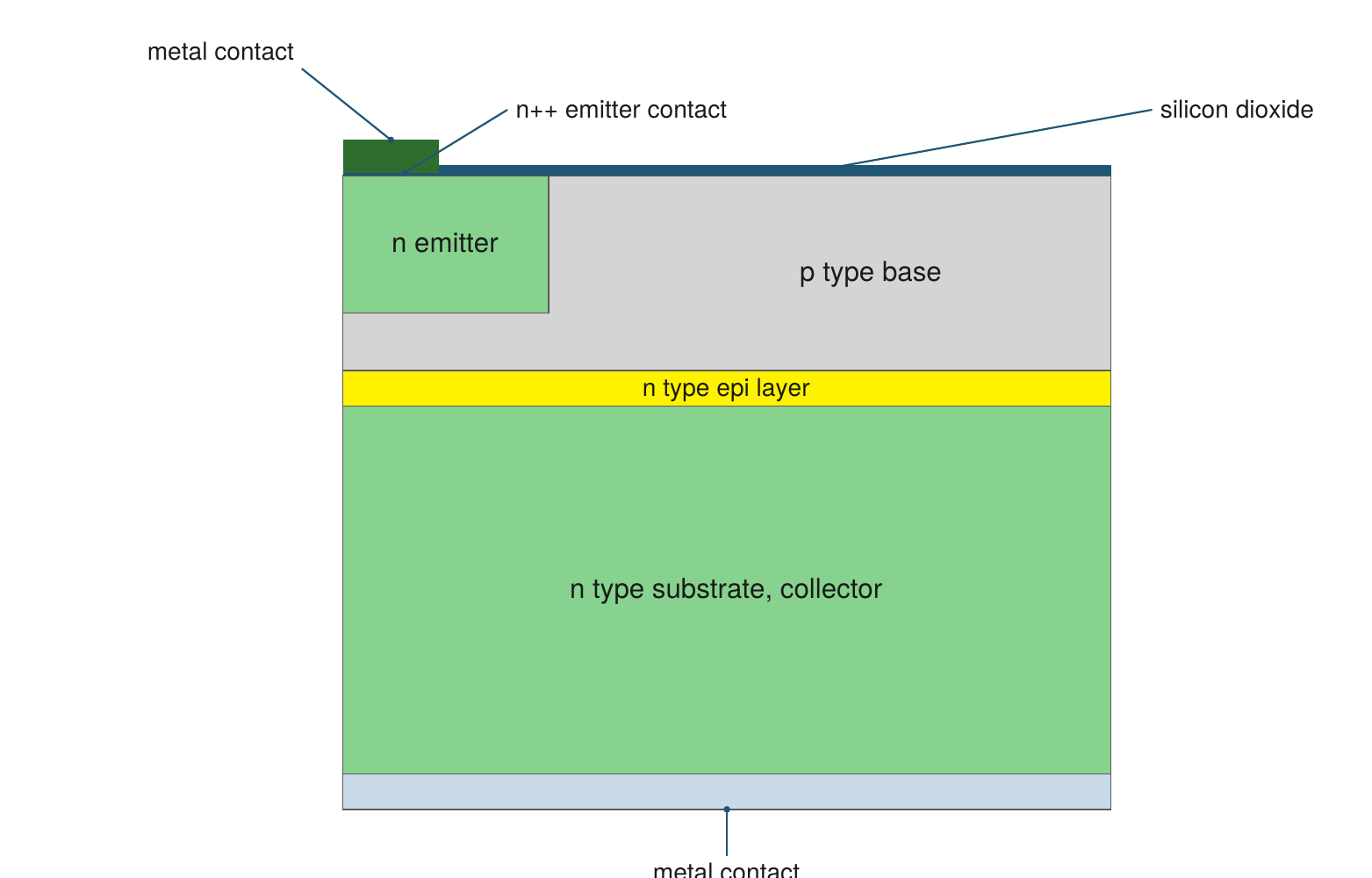}\\[2pt]{\footnotesize (a)}\end{minipage}\hfill
\begin{minipage}{0.32\textwidth}\centering
\includegraphics[width=\linewidth]{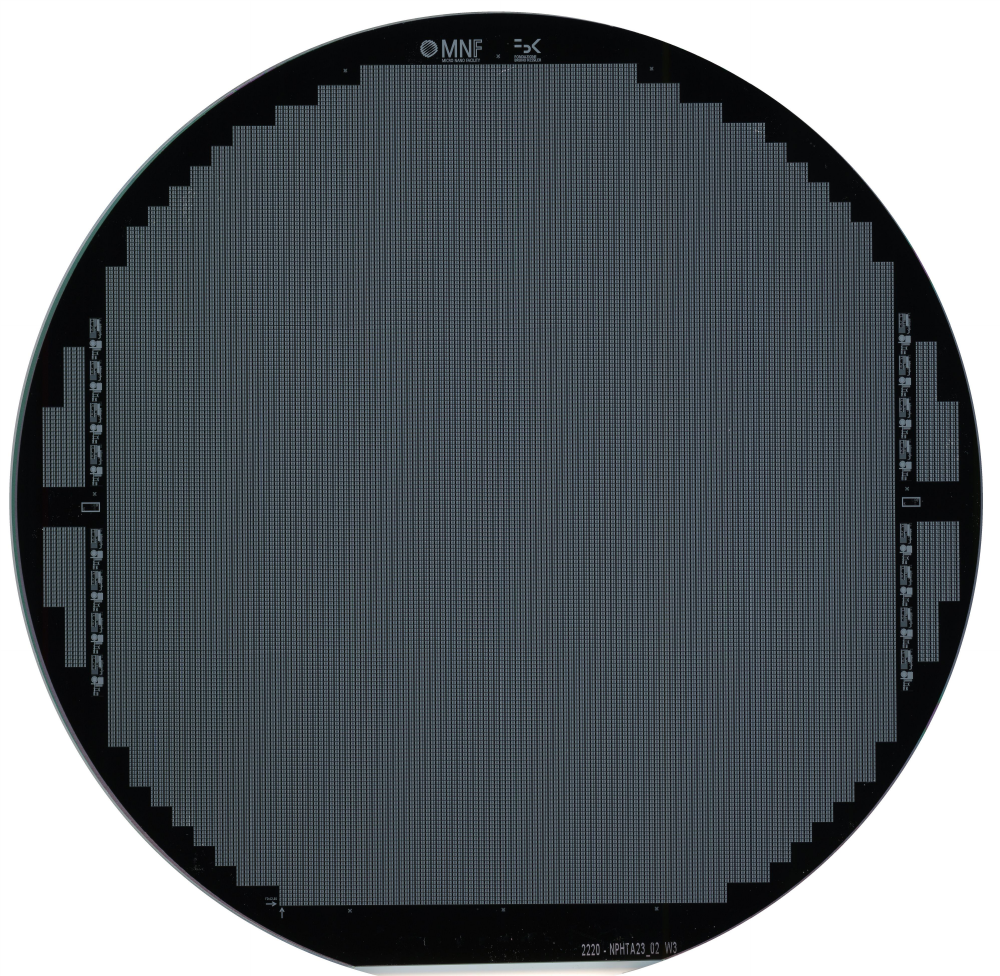}\\[2pt]{\footnotesize (b)}\end{minipage}\hfill
\begin{minipage}{0.32\textwidth}\centering
\includegraphics[width=\linewidth]{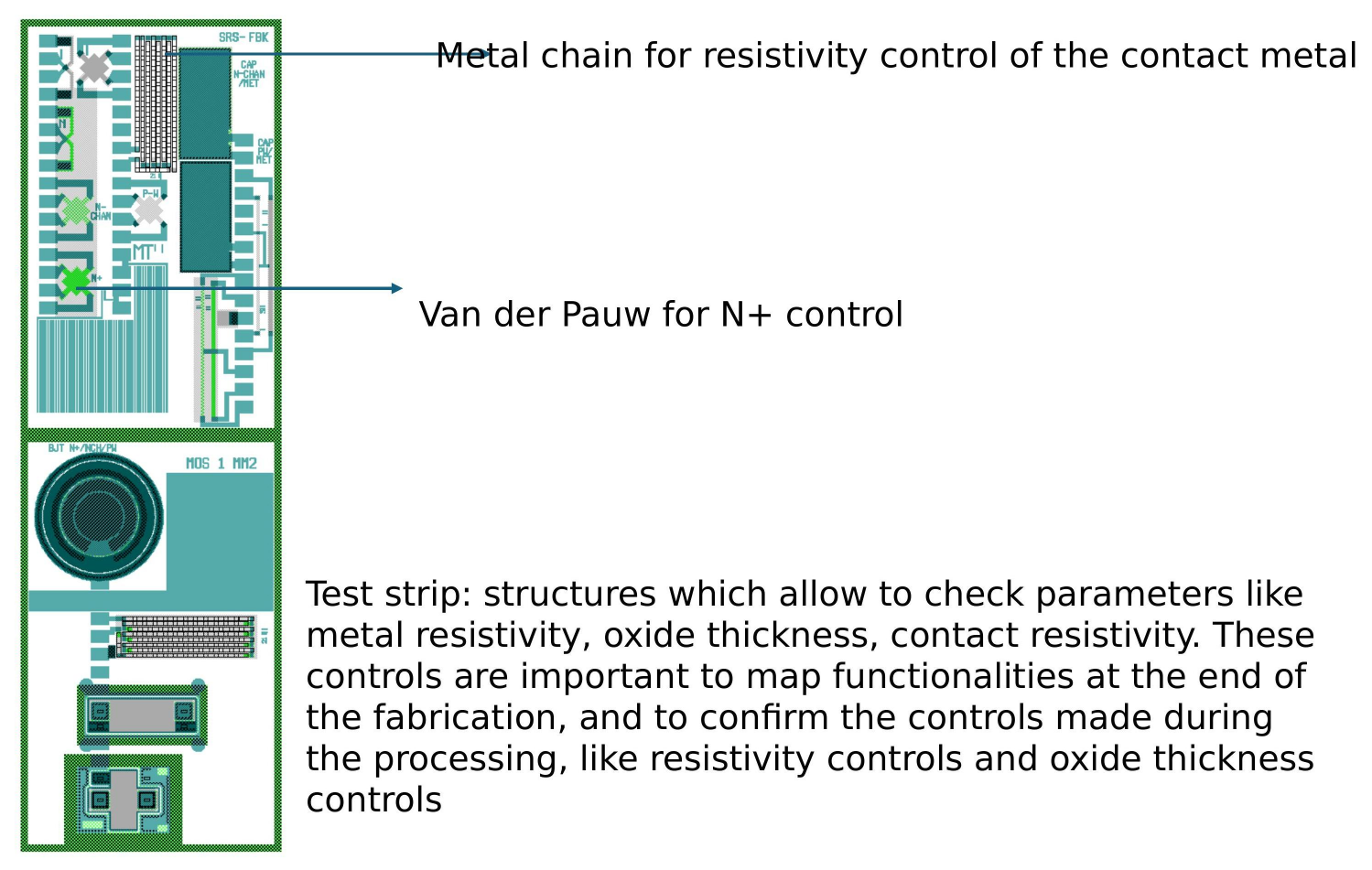}\\[2pt]{\footnotesize (c)}\end{minipage}
\caption{The npn silicon bipolar phototransistor. \textbf{a} Schematic
cross-section showing emitter, base, n-type epitaxial layer, substrate acting as
collector, contacts and screen oxide. \textbf{b} Photograph of a fabricated wafer
carrying the full die array. \textbf{c} On-wafer test-strip structures used to
monitor sheet resistivity, oxide thickness and contact resistivity}
\label{fig:device}
\end{figure}

The device studied here is an npn silicon bipolar phototransistor fabricated by a
planar silicon process, in which the emitter and base are defined by ion
implantation while the collector is provided by an n-type epitaxial layer on the
substrate rather than by implantation. Figure~\ref{fig:device} shows a schematic
cross-section of the device, a fabricated wafer carrying the full die array, and
the on-wafer test structures used to monitor the process. The device descends
from the optical-encoder phototransistor developed at the institute (then IRST,
now Fondazione Bruno Kessler) and optimized through coupled process and device
simulation by \citet{dallabetta1998}, who showed that increasing the emitter
implant dose raises the current gain and used it to bring the gain to
specification. The present work is the data-driven counterpart roughly three
decades on: rather than defining the process by physics-based simulation and
verifying it on fabricated devices, we model gain empirically across many
fabricated wafers.

Fabrication is carried out in a semiconductor cleanroom, where silicon
wafers\footnote{A wafer is a thin, polished disc of crystalline silicon that
serves as the substrate on which many devices are fabricated together, before
being separated (``diced'') into individual chips.} pass through a long sequence
of material depositions, selective removals, thermal and doping steps, and
metrology, interspersed with lithography steps that transfer the device layout
onto the wafer. At the end of fabrication, every wafer is electrically
characterized before it is diced into individual devices.

Because gain is set principally at the emitter, this work focuses on the process
steps that define and control it. The screen oxide is grown first: a layer
(target thickness 100~nm) formed by silicon oxidation in an O$_2$-based furnace
atmosphere, which controls the quantity and depth profile of the implanted
phosphorus and limits crystal damage from the incident ions. Its thickness is
monitored on dedicated test wafers and on some process wafers using an optical
interferometer that maps 16 points per wafer; the mean and standard deviation are
recorded and compared against control limits, with wet-etch thinning or further
oxidation applied where the layer is out of range. Phosphorus is then
ion-implanted to form the n-type emitter and electrically activated by a
high-temperature drive-in (above 1000\,$^{\circ}$C). Finally, the activated
emitter sheet resistivity is measured by a four-point probe on dedicated test
wafers.

\subsection{Process parameters and recipe features}\label{sec:features}

The models use three primary process parameters as recipe features, each
corresponding to one of the emitter-control steps above: the \emph{screen-oxide
thickness}, the \emph{emitter implant dose} and the \emph{emitter sheet
resistivity} after activation. A fourth feature, the \emph{within-batch
resistivity spread}, captures the uniformity of resistivity control.
Table~\ref{tab:features} gives the exact dataset column names, meanings and
units.

All four features are released in normalized form: each is expressed as a ratio
to a per-arm reference (a robust within-arm central value), with the raw values
replaced in place in the published dataset and the reference constants held
internally under the institute's data-sharing policy. This normalization does not
affect any modelling result: each model either is invariant under per-feature
rescaling or is fit within a pipeline that standardizes features on the training
data before fitting, so a fixed linear rescaling of any feature leaves the fitted
predictions unchanged. It is described further in Section~\ref{sec:sharing}.

\begin{table}[htbp]
\centering
\caption{Process parameters and recipe features used as model inputs. All
recipe-defining quantities are released in normalized form}
\label{tab:features}
\setlength{\tabcolsep}{3pt}
\renewcommand{\arraystretch}{1.15}
\begin{tabular}{@{}p{2.1cm}p{2.7cm}p{3.1cm}p{1.5cm}p{1.7cm}@{}}
\toprule
\textbf{Feature (prose name)} & \textbf{Dataset column} & \textbf{Meaning} & \textbf{Units} & \textbf{Form in release} \\
\midrule
Screen-oxide thickness & \texttt{batch\_\allowbreak thickness\_\allowbreak mean\_\allowbreak nm} & Mean screen-oxide thickness (per batch) & nm & Normalized \\
Emitter implant dose & \texttt{emitter\_\allowbreak dose} & Phosphorus implant dose & ions\,cm$^{-2}$ & Normalized \\
Emitter sheet resistivity & \texttt{resistivity\_\allowbreak mean\_\allowbreak ohm\_\allowbreak sq} & Mean activated-emitter sheet resistance & $\Omega$/sq & Normalized \\
Resistivity spread & \texttt{resistivity\_\allowbreak std} & Within-batch resistivity standard deviation & $\Omega$/sq & Normalized \\
\bottomrule
\end{tabular}
\end{table}

\subsection{Gain measurement}\label{sec:gain}

Device current gain $h_{FE}$ is the response variable for all models. After
fabrication, every phototransistor on a wafer is measured to extract its gain,
among other parameters. The wafer-level response used in modelling,
\texttt{gain\_median}, is the median gain over the dies that pass quality
screening, those flagged GOOD and surviving the cleaning steps described in
Section~\ref{sec:dq}, which makes the response robust to die-level outliers and
incomplete measurement coverage.

Gain values are subject to a physical-validity filter of [100, 5000]: readings
outside this window are treated as non-physical and removed before aggregation.
This is an admissibility bound on the measurement, not a yield criterion. The
device-acceptance range desired for the encoder application is [600, 1500]; this
is a design target rather than a data filter, and enters the analysis only as the
target range for the inverse recipe search of Section~\ref{sec:inverse}. It
excludes no wafer from the dataset. Empirically, the modelling corpus spans
[644.0, 1444.5], so the fabricated material sits inside the acceptance window
without that window being imposed.

In addition to the phototransistors, each wafer carries 16 test strips:
monitoring structures placed among the devices that allow parameters such as
metal resistivity, oxide thickness and contact resistivity to be checked at
end-of-line. These independently confirm the in-process thickness and resistivity
controls described in Section~\ref{sec:device-fab}.

\subsection{Dataset structure}\label{sec:dataset}

\begin{figure}[htbp]
\centering
\includegraphics[width=\textwidth]{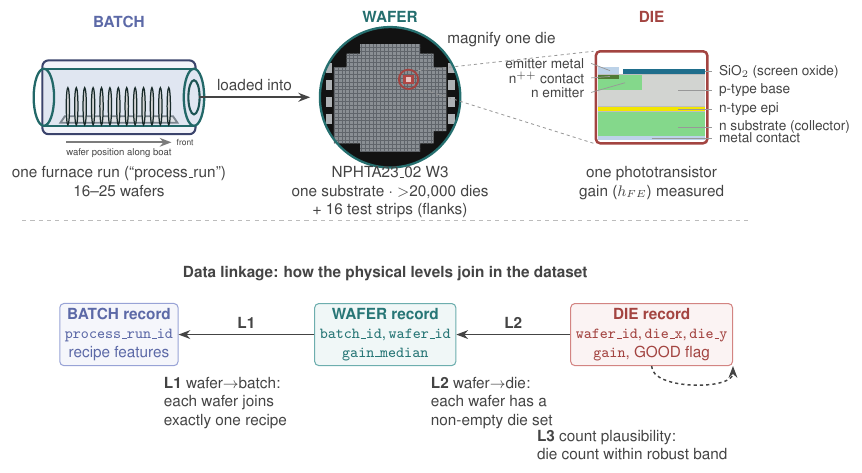}
\caption{The batch/wafer/die hierarchy and its mapping to the dataset. Upper
panel: the physical nesting, in which a furnace run (batch) loads 16--25 wafers
in a quartz boat, each wafer carries more than 20{,}000 dies plus 16 test strips,
and each die is one npn phototransistor whose gain is measured. Lower panel: how
the three physical levels join as dataset records, with the cross-level linkage
checks L1 (wafer-to-batch), L2 (wafer-to-die) and L3 (die-count plausibility)
used by the multi-level data-quality assessment of Section~\ref{sec:dq}}
\label{fig:hierarchy}
\end{figure}

The data are hierarchical, with three strata: fabrication batch, wafer and die.
Thirteen gain-bearing phototransistor batches enter the corpus, each comprising
16 to 25 wafers except batch 3 (45; see below), with more than 20{,}000
phototransistors and 16 test strips per wafer, on the order of 9.3 million
die-level gain measurements in total across the corpus.
Table~\ref{tab:dataset-totals} gives the per-arm totals and
Table~\ref{tab:dataset-runs} the per-run breakdown;
Figure~\ref{fig:hierarchy} shows how the three physical strata nest and how they
join as dataset records.

The modelling unit, however, is the \emph{run} rather than the physical batch,
because one batch must be subdivided. Batch 3 was fabricated as 45 wafers, but a
data-driven analysis (detailed in Sections~\ref{sec:missing}
and~\ref{sec:models}) identifies two sub-populations within it: re-indexing wafer
position at wafer 26 reverses the position--gain correlation, a signature
independent of where the boundary is placed, and a corroborating discontinuity in
median gain appears at the same location. We therefore treat batch 3 as two runs,
labelled b3a (wafers 1--25) and b3b (wafers 26--45), recorded in the data through
a \texttt{process\_run\_id} field that equals the batch number for every batch
except batch 3.

Two analysis arms are then defined, differing by a single documented decision
about one batch. Batch 2 is missing its emitter-resistivity measurement. The
\emph{strict arm} (Arm A) excludes batch 2 entirely, yielding 13 runs over 260
wafers. The \emph{inclusive arm} (Arm B) instead imputes the missing resistivity,
on the reasoning that batch 2 retains valid thickness, gain and other data, and
need not be discarded over a single absent feature, yielding 14 runs over 285
wafers. The 25-wafer difference between the arms is exactly the batch-2 cohort.
We carry both arms through the modelling and treat the comparison between them as
a sensitivity analysis on the imputation decision; whether including the imputed
batch changes any conclusion is reported in Section~\ref{sec:models}. The two
arms share roughly 91\% of their wafers, so their agreement is a check on the
single imputation decision rather than independent validation; we do not treat
arm agreement as confirmation of a result by independent data. Note that run
counts of 13 and 14 correspond to 12 and 13 distinct batch identifiers
respectively, since the single physical batch 3 contributes two runs in each arm.

\begin{table}[htbp]
\centering
\caption{Dataset summary: per-arm totals. Arm A is the strict, primary arm; Arm B
is the inclusive arm used for sensitivity analysis}
\label{tab:dataset-totals}
\begin{tabular}{lrrrr}
\toprule
Arm & Batches & Process runs & Wafers & Die measurements \\
\midrule
A, strict (primary) & 12 & 13 & 260 & 8{,}597{,}910 \\
B, inclusive (sensitivity) & 13 & 14 & 285 & 9{,}292{,}910 \\
\bottomrule
\end{tabular}
\end{table}

\begin{table}[htbp]
\centering
\caption{Per-run breakdown of the corpus, showing the batch-3 split into runs 3a
and 3b and the batch-2 asymmetry between the two arms. A check mark indicates
that the run is included in that arm}
\label{tab:dataset-runs}
\begin{tabular}{llrcc}
\toprule
Process run & Physical batch & Wafers & Arm A & Arm B \\
\midrule
1 & 1 & 23 & \checkmark & \checkmark \\
2 & 2 & 25 & & \checkmark \\
3a & 3 & 25 & \checkmark & \checkmark \\
3b & 3 & 20 & \checkmark & \checkmark \\
4 & 4 & 20 & \checkmark & \checkmark \\
5 & 5 & 16 & \checkmark & \checkmark \\
6 & 6 & 20 & \checkmark & \checkmark \\
7 & 7 & 20 & \checkmark & \checkmark \\
8 & 8 & 20 & \checkmark & \checkmark \\
9 & 9 & 20 & \checkmark & \checkmark \\
10 & 10 & 20 & \checkmark & \checkmark \\
11 & 11 & 20 & \checkmark & \checkmark \\
12 & 12 & 16 & \checkmark & \checkmark \\
13 & 13 & 20 & \checkmark & \checkmark \\
\bottomrule
\end{tabular}
\end{table}

\section{Data quality}\label{sec:dq}

The models in Section~\ref{sec:models} predict device gain from a small set of
process parameters that vary at the nanoscale: implant doses, an oxide thickness
with a 100~nm target, a sheet resistivity. At that scale small data defects
translate into real predictive error, since a single mis-keyed wafer, an
unflagged measurement artefact, or a hidden change in process conditions can move
a fitted coefficient more than the physical effect it is meant to capture. A
large empirical study of how data-quality dimensions affect machine-learning
performance across nineteen algorithms finds that completeness, feature accuracy
and target accuracy degrade supervised models the most, while consistency and
uniqueness matter comparatively little; and, relevant to the small,
high-feature-count corpus here, that smaller datasets and those with many
features relative to their sample size showed greater sensitivity along several
of these dimensions, with the largest dataset consistently the most robust
\citep{mohammed2025}. For a small-sample fabrication corpus, then, the
trustworthiness of the data is not a hygiene step but a precondition for the
modelling. This section establishes that the dataset the models rest on is
trustworthy, how missing and anomalous values were handled, and how the release
is structured so the assessment can be repeated as future batches are added.

\subsection{Why hierarchical data-quality assessment is needed for fabrication data}\label{sec:why-hier}

The foundational dimensional vocabulary of \citet{wangstrong1996}, recent
surveys of data-quality dimensions and tooling for machine learning
\citep{zhou2024}, and hierarchical consolidations of that vocabulary such as the
ISO/IEC~25012 extension of \citet{miller2024} all specify their dimensions for a
single, undifferentiated dataset, with no mechanism for scoring distinct nested
levels or the correspondence between them. Where such frameworks are themselves
called hierarchical, the hierarchy organizes dimension \emph{categories}
(intrinsic, contextual, representational, and so on) rather than levels of the
data. Fabrication data does not fit that shape. It is organized in strict
physical strata: a batch is processed under one recipe, a wafer is one substrate
processed within that batch, and a die is one measured device, and the same
quantity behaves differently at each stratum. A recipe parameter is one value per
batch but broadcasts across twenty-odd wafers on joining, while gain exists only
at the die level and is a summary everywhere above. A single-level rubric must
therefore either pick one level and discard the rest, or flatten the levels and
double-count the broadcast values.

Fabrication data needs both of the pieces identified as missing in
Section~\ref{sec:rw-dq}, because the failures that matter most are cross-level
integrity failures (a wafer tracing to no
recipe, an implausible die count, a batch identifier concealing two process
runs) that are properties of the \emph{links} between levels. The corpus is also
small where modelling decisions are made (twelve to thirteen batches), so
distribution-dependent dimensions are meaningless at the batch level yet
well-defined at the die level, where measurements number in the millions. The
multi-level data-quality assessment (MLDQ) developed here responds to both
constraints: it extends per-level scoring to the nested \emph{physical} entities
of fabrication data, with a dimension set tailored to each level, and adds an
explicit cross-level linkage score as a first-class dimension.

\subsection{MLDQ methodology}\label{sec:mldq-method}

MLDQ scores nine data-quality dimensions, summarized in
Table~\ref{tab:mldq-dims}: Traceability (records carrying batch and wafer
identifiers), Completeness (non-missing fraction over in-scope columns), Accuracy
(values within physical hard bounds), Consistency (column-name and unit
conformance), Redundancy (penalizing near-duplicate columns at $|r| > 0.95$),
Balance (gain-distribution shape and good-die ratio), Uniqueness (non-duplicate
records on the natural key), Usefulness (leave-one-batch-out\footnote{Leave-one-batch-out
cross-validation (LOBO-CV): a validation scheme in which one entire batch
(process run) is withheld as the test set while the model is trained on all
remaining batches, repeated so that every batch is held out exactly once; unlike a
random train--test split it measures how well the model predicts a run it has
never seen.} predictive $R^{2}$ of a wafer-level model), and Quantity
(sample-to-feature ratio). The dimension set is rooted in the intrinsic and
contextual data-quality tradition established by \citet{wangstrong1996}. Most
dimensions are taken from the materials-data governance framework of
\citet{liu2025a} (Traceability, Completeness, Consistency, Redundancy and
Balance). Usefulness and Quantity follow the ISO/IEC~25012 extension of
\citet{miller2024}, with the sample-to-feature framing of Quantity also drawing on
\citet{liu2023}; Accuracy follows the single-descriptor rule of \citet{liu2025b}
detailed below; and Uniqueness is a standard referential-integrity dimension. The
operational definitions used here, among them the $|r|>0.95$ redundancy threshold
and the leave-one-batch-out $R^{2}$ used for Usefulness, are specific to MLDQ
rather than inherited from these sources. The Accuracy dimension builds on Rule~1
(the value-and-grammar rule) of the single-descriptor accuracy detection model
(S-DAD) of \citet{liu2025b}, which checks each descriptor value against a declared
empirical range together with type and unit conformance. We adopt that symbolic
range rule alone, encoding the bounds in a machine-readable rules file; we
deliberately do not apply the statistical-detector ensemble that S-DAD intersects
with the rule. At this sample size, distribution-based detectors flag physically
valid extremes as anomalies (a false-positive mode the S-DAD authors document on
small, high-diversity data), and, because a value can violate a hard physical
bound while still falling within the bulk of an otherwise unremarkable
distribution, the range rule on its own is the more reliable guard against
physically impossible values here. Timeliness is omitted throughout, as the data
are static.

The contribution is not running this rubric three times but deciding which
dimensions are \emph{meaningful} at which granularity, and stating which are not
and why. Each level scores a tailored subset (Table~\ref{tab:mldq-dims}), and
every exclusion carries a reason rather than a silent omission. At the
\emph{batch} level (recipe space, $N = 12$--$13$), Quantity applies \emph{as a
deliberately low score}: against roughly thirty features it scores around ten
(10.2 in the baseline and Arm B, 9.4 in Arm A). We treat that low score as
information rather than a defect to suppress:
it is the quantitative signal that batch-level modelling is ill-supported.
Reviews of small-sample materials machine learning observe that the
feature-to-sample ratio, rather than feature count or sample count alone, is the
quantity that governs this regime, reporting that the imbalance persists even when
samples outnumber features roughly four-fold \citep{liu2023}. Rather than respond
with the synthetic augmentation those reviews catalog among the standard
remedies, which for physics-grounded measurements would fabricate process
conditions never observed in the fab, MLDQ surfaces the scarcity as a scored
quantity and lets it justify two adaptations downstream: no synthetic
augmentation, and analysis shifted to the denser wafer level under the
hierarchical models of Section~\ref{sec:models} rather than recipe-only fitting at
the batch level. The batch-level exclusions each carry a reason: Redundancy
because Pearson correlation over a dozen rows is meaningless, Balance because
distribution shape is a die-level property, and Usefulness because the recipe-only
model does not generalize across runs, its leave-one-batch-out $R^{2}$ being at
best marginal (itself a scientific finding, reported
in Section~\ref{sec:ridge}); Uniqueness is inherited, batch identity being the
trivially unique key. At the \emph{wafer} level ($N \approx 285$) all nine apply,
with one coherence fix: recipe-origin columns are scored once at batch level and
inherited at wafer level, so a single missing recipe is not counted as missing
across the twenty wafers it broadcasts to (the W1 inheritance rule). At the
\emph{die} level (millions of rows) the per-wafer scores aggregate to a
distribution rather than a mean, surfacing outlier wafers. Die-level Accuracy uses
the \emph{physical} gain bound [100, 5000], an out-of-range value being an
acquisition glitch, deliberately not the engineering acceptance range [600, 1500],
which is a yield metric never folded into a quality score. Each level reports the
mean of its scored dimensions only, so the level means are comparable without
inflation by trivial perfect scores. The exclusion-with-reason pattern is itself
part of the contribution: it is the visible evidence the design is tailored to
each stratum.

The cross-level \emph{linkage} dimension is the new scored quantity, measuring how
cleanly the hierarchy joins through three sub-scores: L1 (wafer-to-batch), the
fraction of wafers joining exactly one recipe; L2 (wafer-to-die), the fraction
with a non-empty die set; and L3 (count plausibility), the fraction whose die
count falls within a robust band around the corpus median. The combined
cross-level linkage score is a weighted mean of the three sub-scores,
\begin{equation}
\mathrm{Linkage} = 0.40\,L_1 + 0.40\,L_2 + 0.20\,L_3,
\label{eq:linkage}
\end{equation}
where $L_1$ is the wafer-to-batch referential integrity, $L_2$ the wafer-to-die
completeness, and $L_3$ the die-count plausibility: the two hard
structural-integrity checks carry equal high weight, while L3 is soft and carries
less, since an unusual count may be valid. The L3 plausibility band around the
median die count $m$ is
\begin{equation}
\text{band} = m \pm \max\!\left(3\,\mathrm{MAD},\; 0.05\,m\right),
\label{eq:band}
\end{equation}
where $\mathrm{MAD}$ is the median absolute deviation of the per-wafer die counts
and the five-percent-of-median floor prevents the band from collapsing when the
MAD is zero, as occurs here (Section~\ref{sec:mldq-applied}); using median and MAD
rather than mean and standard deviation follows standard robust practice
\citep{leys2013}. All three sub-scores are reported alongside the combined number,
and linkage is never folded into a level mean.

\begin{table}[htbp]
\centering
\caption{MLDQ dimensions, sources and per-level applicability. In the three
right-hand columns, A denotes a dimension that applies and is scored, INH a
dimension that is inherited or structural and is therefore reported but excluded
from the level mean, and X a dimension excluded for the stated reason. Each level
mean is taken over its A-dimensions only, and Timeliness is omitted globally
because the data are static}
\label{tab:mldq-dims}
\setlength{\tabcolsep}{3pt}
\renewcommand{\arraystretch}{1.15}
\begin{tabular}{@{}p{2.0cm}p{1.9cm}p{3.5cm}p{1.5cm}p{1.0cm}p{1.5cm}@{}}
\toprule
\textbf{Dimension} & \textbf{Source} & \textbf{Scoring method} & \textbf{Batch} & \textbf{Wafer} & \textbf{Die} \\
\midrule
Traceability & \citet{liu2025a} & \% records with batch\_id and wafer\_id & A & A & INH \\
Completeness & \citet{liu2025a} & mean non-missing fraction, in-scope columns & A & A & A \\
Accuracy & \citet{liu2025b}, S-DAD Rule 1 only & $100\cdot(1 - \text{violations/total})$ against hard bounds & A & A & A (physical) \\
Consistency & \citet{liu2025a} & column-name and unit conformance & A & A & X (scored once) \\
Redundancy & \citet{liu2025a} & $100\cdot(1 - \text{redundant pairs/total})$, $|r|>0.95$ & X ($N$=12--13, noise) & A & X (signal) \\
Balance & \citet{liu2025a} & gain skewness and good-die ratio & X (die-level) & A & A \\
Uniqueness & standard; cf.\ \citet{miller2024} & \% unique on natural key & INH & A & A \\
Usefulness & \citet{miller2024} & LOBO-CV ridge $R^{2}$ & X (negative $R^{2}$ is a finding) & A & X (no die model) \\
Quantity & \citet{miller2024}; \citet{liu2023} & $\min(1, N/(4\cdot\text{features}))\cdot100$ & A (deliberately low) & A & X (trivially 100) \\
Linkage & novel & $0.40L_1 + 0.40L_2 + 0.20L_3$ & \multicolumn{3}{c}{cross-level} \\
\bottomrule
\end{tabular}
\end{table}

\subsection{Handling missing and anomalous data: two-arm imputation and the batch-3 split}\label{sec:missing}

Two data-handling decisions shape the modelling corpus, and rather than commit to
one treatment of each, we carry the alternatives as explicit sensitivity
questions, the principle being that with so few batches a single discretionary
choice should never silently determine a result.

The first is missing data. Batch 2's sheet resistivity failed measurement, and
batch 8 carries a structural gap in its resistivity spread from a single-spot
protocol. The two arms defined in Section~\ref{sec:dataset} differ exactly in the
treatment of batch 2. Batch 8's missing resistivity spread is filled in both arms, as it is the only
resistivity-spread gap and does not depend on the batch-2 decision; the
pre-imputation baseline scope leaves it unfilled. Imputed values are unweighted
donor-pool means over the batches with a measured value, unweighted at batch level
so larger batches cannot dominate, and every imputed cell carries a row-level
provenance flag. After publication normalization, the imputed resistivity mean
coincides with the corpus central value (a normalized ratio of one), and the
imputed spread corresponds to normalized ratios of about 1.11 (Arm A) and 1.07
(Arm B). Because both imputed quantities are derived summary statistics rather
than raw measurements, this gap-filling is categorically distinct from the
synthetic data augmentation the framework declines. Crucially, we do not claim the
imputation is harmless: the resistivity spread is a predictively relevant feature
in the models of Section~\ref{sec:models}, so rather than assume its fill leaves
conclusions untouched, we carry both arms and test that directly. As
Section~\ref{sec:mldq-applied} shows, the quality scores themselves are by
construction invariant to the imputation value, responding only to the more
consequential decision of whether the batch is retained; whether any modelling
conclusion shifts is reported in Section~\ref{sec:models}.

\begin{figure}[htbp]
\centering
\includegraphics[width=0.8\textwidth]{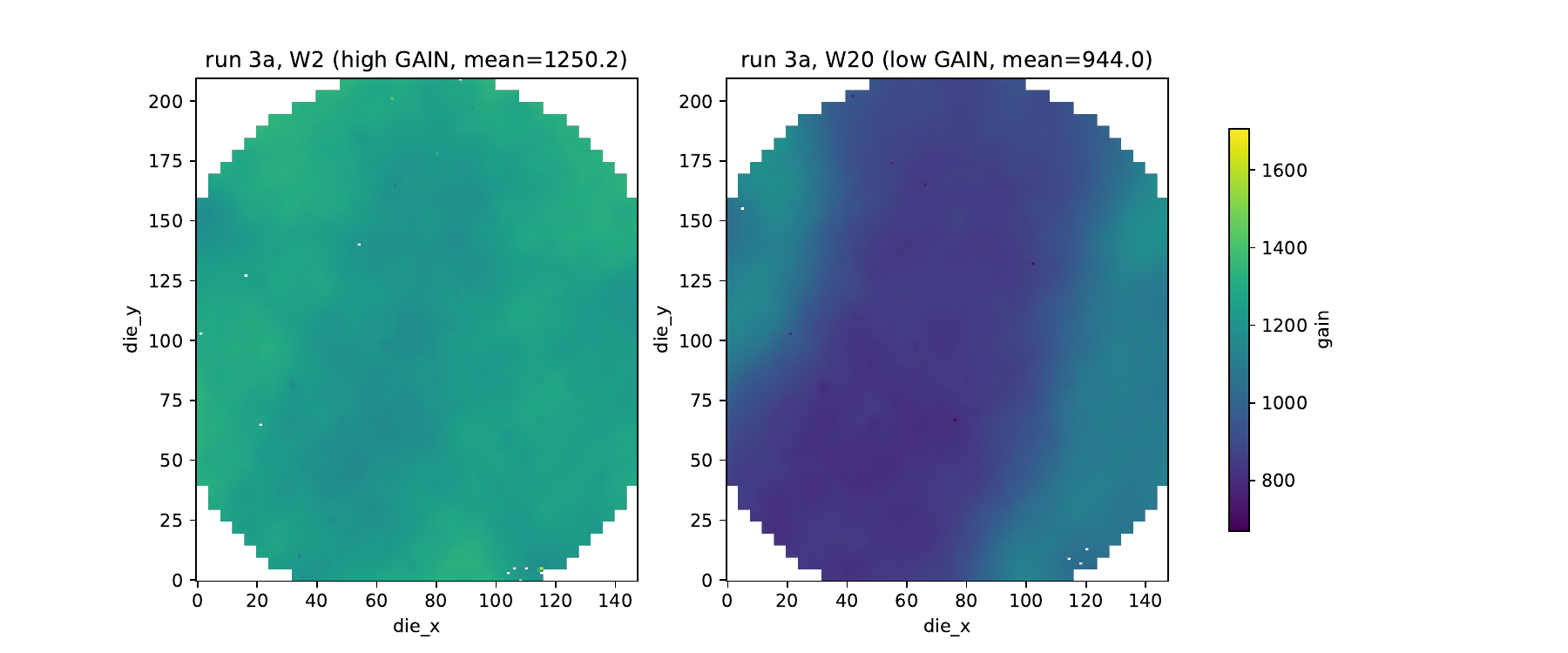}
\caption{Die-level gain map for process run 3a, showing the spatial structure of
gain across the wafer array}
\label{fig:spatial_map}
\end{figure}

The second is anomalous structure, and it was a discovery rather than a decision.
Batch 3 contained 45 wafers, twice the corpus norm, and examining that anomaly
revealed two independent lines of evidence for two sub-populations. The stronger,
and the one our split rests on, is a reversal in the position effect: a Spearman
correlation between wafer position and gain of $+0.25$ over all 45 wafers becomes
$-0.47$ ($p = 0.001$) once position is re-indexed from the 26th wafer, with a
coherent negative within-segment slope in each half (b3b within-half $\rho =
-0.50$, $p = 0.025$). This re-indexing signature does not depend on where the
boundary is placed and is therefore unaffected by the boundary having been
selected from a scan. A second, corroborating signature is a discontinuity in
median gain of about 80 units at the same location. We report this gap with
appropriate caution: because the W25/W26 boundary was selected as the sharpest of
36 candidate cuts (W5 to W40), the naive two-sample $p$-value (Welch $t = -3.35$,
$p = 1.8 \times 10^{-3}$) overstates the evidence. A permutation test over the
maximized discontinuity statistic, which corrects for the selection, gives $p =
0.16$; a gap of this size arises in roughly one in six datasets of this length
under the null of no change-point. The gap is therefore consistent with the split
but, on its own, not significant once the search is accounted for. The split is
justified by the selection-independent re-indexing signature, with the gap as
corroboration. We therefore treat batch 3 as two consecutive process runs under
one identifier (25 wafers then 20), encoded as a process-run label that leaves
every source measurement unaltered. All results reported below use this split. The most plausible physical reading is that a
45-wafer batch spans two loading cassettes, since a standard cassette holds about
25 wafers, and the discontinuity falls where that boundary would lie; direct
cassette records are not available, so this is offered as the likeliest
explanation rather than a confirmed cause, and the statistical evidence for the
split stands independently of it. Primary results are reported with the split, and
the corresponding results without it are given in Online Resource~1.
Figure~\ref{fig:spatial_map} shows the die-level gain map for the first of the two
resulting runs. The discovery is itself an argument for hierarchical assessment:
one batch identifier concealed two physically distinct populations, and comparing
structure across levels exposed what a flat assessment of the wafer table would
not. The full modelling consequence of the split is taken up in
Section~\ref{sec:models}.

\subsection{Data sharing, normalization and reuse}\label{sec:sharing}

The dataset is released to support reuse while respecting the confidentiality of
the underlying recipes, and is structured so the quality assessment can be re-run
as new batches arrive. The public release centres on the wafer-level table,
per-wafer gain summaries, the derived process-run identifier, and the recipe and
substrate features used for modelling, accompanied by the quality reports, the
modelling outputs and figures, the analysis scripts, and a machine-readable schema
documenting every column, its role and its validation bounds. The raw die-level
records are retained internally: they number in the millions, are the most
sensitive layer, and are not required to reproduce any result here, since all
modelling operates on wafer-level summaries. Because the schema encodes each
column's validation bounds and the pipeline is runnable end to end, appending a
future batch re-runs the same per-level scores and linkage checks without rework;
the assessment is designed to be repeated, not performed once.

Recipe-defining quantities are normalized before release: absolute implant doses,
resistivities and oxide thicknesses are replaced by values expressed relative to
internally held references, preserving the relative structure the models depend
on (orderings, ratios and relationships among parameters) while withholding the
absolute specification. Every public-pipeline analysis runs identically on the
normalized data, and the modelling results are numerically identical between the
internal and published representations to within floating-point tolerance: the
mixed-effects and Gaussian-process models are invariant under per-feature
rescaling, and the ridge and random-forest models are fit within a pipeline that
standardizes features on the training fold before fitting, so per-feature
rescaling leaves their predictions unchanged. Normalization
therefore changes the numerical representation of the recipe features without
changing any modelling result. The normalized wafer-level dataset is deposited on
Zenodo under a CC-BY-4.0 licence with a citable digital object identifier
\citep{amirabgir2026data}, and the analysis code, trained models and the same
normalized dataset are released together on GitHub, so every result here is
reproducible from the public artefacts without access to internal data; the raw
die-level records and the normalization constants remain internal. The release is
organized along the FAIR (findable, accessible, interoperable, reusable)
principles \citep{wilkinson2016}: \emph{findable} through a persistent identifier
and deposition in a searchable, indexed repository; \emph{accessible} through that
identifier over a standard, open protocol; \emph{interoperable} through a
documented, machine-readable schema; and \emph{reusable} under the stated licence,
with recorded provenance and per-column validation bounds.

\subsection{MLDQ applied: scoring matrix, robustness and what linkage surfaces}\label{sec:mldq-applied}

Applying MLDQ to the cleaned corpus produces a per-level scoring matrix across the
three strata plus the linkage dimension, evaluated under three scopes: a
pre-imputation baseline (285 wafers, batch 3 scored as one batch-level row, batch
2 carrying missing resistivity), Arm A (260 wafers, batch 2 dropped) and Arm B
(285 wafers, batch 2 imputed). Comparing the matrix across scopes is itself a test
of the method: a scheme whose verdicts swung with an arguable imputation choice
would have limited diagnostic value.

The headline result is that MLDQ is \emph{robust to the imputation choice by
construction}. The baseline and Arm B scopes produce identical scores at every
level mean and all nine wafer-level dimensions, to three decimal places, because
of the W1 inheritance rule: batch 2's resistivity, missing in the baseline and
imputed in Arm B, is scored at batch level and never enters the wafer-level
Completeness numerator, so the single imputed value cannot move a wafer-level
score. The only divergence comes from Arm A's stricter rule, which removes batch 2
entirely, and even that stays under half a point at every level (batch mean
$+0.06$, wafer $-0.30$, die $+0.31$, combined linkage $-0.20$;
Table~\ref{tab:mldq-scopes}). The batch level shows why this is stability rather
than mere insensitivity: dropping batch 2 lowers Quantity by roughly 0.8 points
(smaller $N$) but raises Accuracy by roughly the same amount, because batch 2's
unmeasured resistivity had violated the hard minimum and that violation leaves the
scored set when the batch is dropped, so the two effects very nearly cancel. The
framework thus surfaces the consequential drop-versus-keep decision while
remaining invisible to the imputation value.

\begin{figure}[htbp]
\centering
\includegraphics[width=0.92\textwidth]{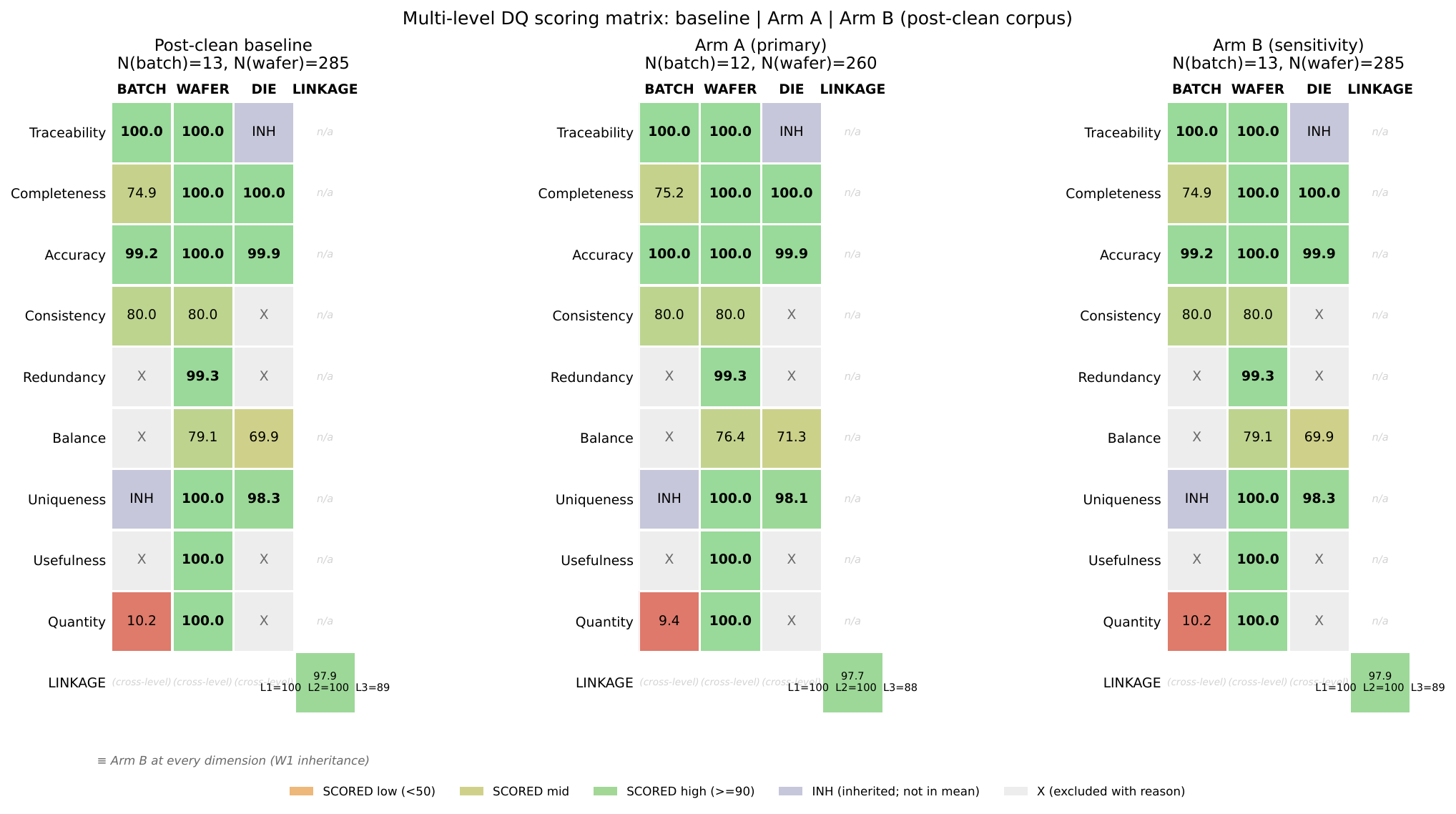}
\caption{MLDQ scoring matrix under three corpus scopes: baseline, Arm A and Arm B.
Each panel gives the per-dimension scores at the batch, wafer and die levels
together with the linkage sub-scores. The baseline and Arm B panels are identical
by the inheritance rule, and the Arm A differences are at most half a point at
every level mean}
\label{fig:mldq-matrix}
\end{figure}

The level means read against design intent, not as raw grades. The batch mean of
72.9 is low by design, dragged by the Quantity score that encodes small $N$ as a
measurable property; the wafer mean of about 95 is the canonical
``is the dataset usable'' figure; and the linkage score of about 98 is high
because L1 and L2 are both perfect, with only L3 below 100.
Figure~\ref{fig:mldq-matrix} presents the full matrix as three heatmap panels, the
baseline and Arm B panels being identical by the W1 rule.

That L3 score is where linkage earns its place. L3 scores about 89 in both arms;
the per-wafer die-count distribution is so concentrated at its modal value that
the MAD is zero, the $3\,\mathrm{MAD}$ term in Eq.~\eqref{eq:band} vanishes, and
the five-percent-of-median floor becomes the active band. Without that floor the
band would collapse to a single value and score every routine variation as
anomalous. What L3 surfaces is not one anomaly but three structurally distinct
patterns, all in batches 5, 8 and 11: a deliberate double-measurement protocol at
exactly twice the modal count; extra measurement-letter scans 14--23\% above
modal; and a small set of extreme overcounts several times modal. These have three
different causes (a measurement artefact, an intentional protocol and a genuine
overcount), yet a single-level wafer scorer would have absorbed all of them
without comment, since each wafer's own gain summary is internally complete and
consistent. Only the cross-level check, comparing each die count against the
corpus envelope, flags them and separates them into meaningful groups. Linkage
earns its place not by lowering the score but by surfacing structure the per-level
dimensions cannot see. Online Resource~1 plots each wafer's die count against its
batch with the band overlaid, showing the three tiers as distinct clusters above
the dense in-band population.

\begin{table}[htbp]
\centering
\caption{Three-scope MLDQ comparison. The baseline and Arm B scopes are identical
by the inheritance rule; only Arm A, which drops batch 2, diverges, and by at most
half a point at any level mean. The two right-hand columns give the difference of
each arm from the baseline. The batch, wafer and die level means are computed over
13, 285 and 285 records respectively in the baseline and Arm B scopes, and over
12, 260 and 260 records in Arm A}
\label{tab:mldq-scopes}
\setlength{\tabcolsep}{4pt}
\renewcommand{\arraystretch}{1.15}
\begin{tabular}{@{}lrrrrr@{}}
\toprule
\textbf{Level or dimension} & \textbf{Baseline} & \textbf{Arm A} & \textbf{Arm B} & \textbf{$\Delta$A} & \textbf{$\Delta$B} \\
\midrule
Batch level mean & 72.86 & 72.92 & 72.86 & $+0.06$ & $+0.00$ \\
Wafer level mean & 95.37 & 95.07 & 95.37 & $-0.30$ & $+0.00$ \\
Die level mean & 92.02 & 92.33 & 92.02 & $+0.31$ & $+0.00$ \\
Linkage, combined & 97.89 & 97.69 & 97.89 & $-0.20$ & $+0.00$ \\
L1 wafer$\rightarrow$batch & 100.00 & 100.00 & 100.00 & $+0.00$ & $+0.00$ \\
L2 wafer$\rightarrow$die & 100.00 & 100.00 & 100.00 & $+0.00$ & $+0.00$ \\
L3 count plausibility & 89.47 & 88.46 & 89.47 & $-1.01$ & $+0.00$ \\
\bottomrule
\end{tabular}
\end{table}

\section{Models and findings}\label{sec:models}

This section presents the modelling in two parts. The first
(Sections~\ref{sec:variance}--\ref{sec:ridge}) is the scientific finding: device
gain is governed as much by between-run variance as by the recipe, and the
within-run position effect changes sign from run to run, a structure invisible to
any single pooled model. The second
(Sections~\ref{sec:forward}--\ref{sec:selection}) is the prediction capability:
forward gain prediction, inverse recipe search and a relative uncertainty signal.
Throughout, the strict arm (Arm A, which imputes nothing) is the primary reference
and the inclusive arm (Arm B) is reported as a sensitivity check; the two agree on
every qualitative conclusion, and where they differ numerically the difference is
small and stated.

\subsection{Variance hierarchy: why a recipe-only model cannot succeed}\label{sec:variance}

\begin{figure}[htbp]
\centering
\includegraphics[width=0.8\textwidth]{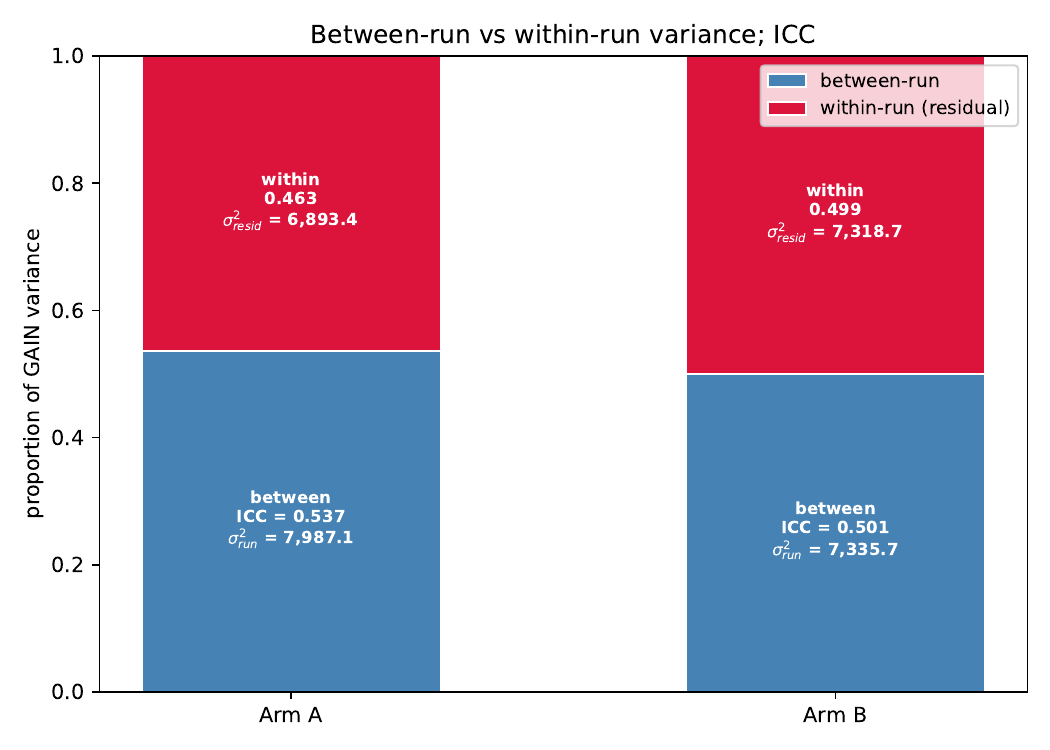}
\caption{Variance decomposition of wafer-level gain into between-run and
within-run (residual) components for Arm A and Arm B, with the intraclass
correlation annotated on each panel. Roughly half the variance lies between runs
in both arms}
\label{fig:variance}
\end{figure}

\begin{figure}[htbp]
\centering
\includegraphics[width=0.7\textwidth]{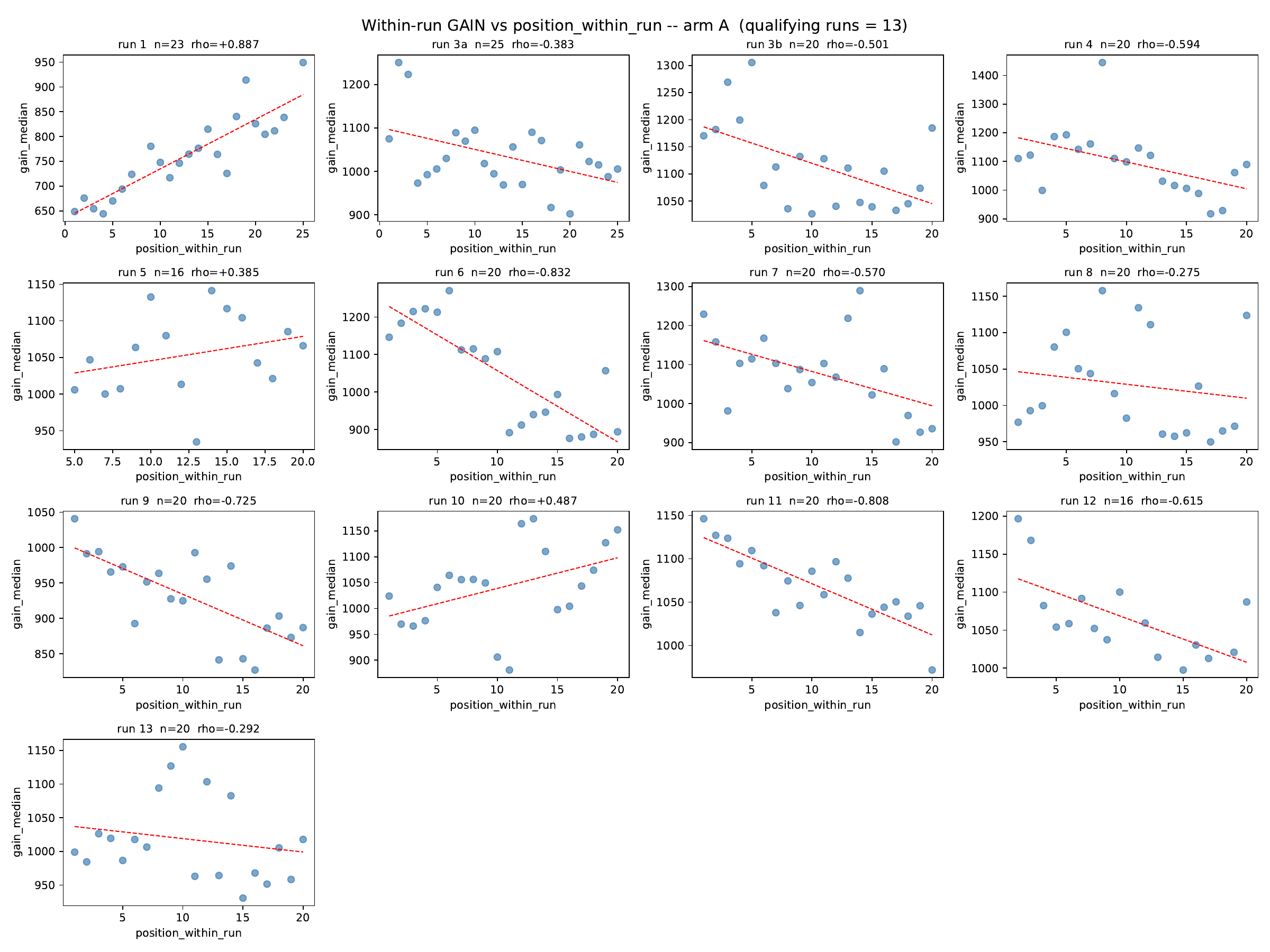}
\caption{Within-run gain spread across process runs for Arm A, complementing the
variance decomposition by showing the residual within-run variability that
recipe-only models cannot access}
\label{fig:within_run_spread}
\end{figure}

Before fitting any predictive model we decompose the variance of wafer-level gain
into between-run and within-run components, because the result determines which
model classes can work at all. A one-way decomposition with the process run as the
grouping factor gives an intraclass correlation coefficient (ICC)
\citep{nakagawa2017}, the share of total variance attributable to the run,
\begin{equation}
\mathrm{ICC} = \frac{\sigma^2_{\text{run}}}{\sigma^2_{\text{run}} + \sigma^2_{\text{resid}}},
\label{eq:icc}
\end{equation}
where $\sigma^2_{\text{run}}$ is the between-run variance and
$\sigma^2_{\text{resid}}$ the within-run (residual) variance. In our corpus this
ICC is 0.537 in Arm A and 0.501 in Arm B; that is, roughly half of the total
variance in gain lies \emph{between} runs rather than within them at the point
estimate. Figure~\ref{fig:variance} shows the decomposition for both arms, and
Figure~\ref{fig:within_run_spread} the residual within-run variability that
recipe-only models cannot access. The between-run and residual variance components
are comparable in magnitude (Arm A: between-run 7987, residual 6893; Arm B: 7336
and 7319 in the same standardized units), and an independent estimate from the
null mixed-effects model agrees (ICC 0.516 and 0.481 respectively). The
corresponding one-way analysis of variance rejects equality of run means in both
arms ($F = 26.4$, $p \approx 10^{-38}$ in Arm A; $F = 22.9$, $p \approx 10^{-36}$
in Arm B).

A cluster bootstrap over whole process runs ($B = 2000$) places the 95\% interval
at [0.03, 0.74] (Arm A) and [0.03, 0.73] (Arm B), with 65\% and 63\% of resamples
respectively retaining an ICC above 0.40.\footnote{Variance-component estimates
are sensitive to the fitting method at this sample size. A single-start L-BFGS
REML fit converges to the zero-variance boundary in 41\% (Arm A) and 49\% (Arm B)
of resamples, which depresses the lower confidence bound to zero; in simulation
with the same group structure and a known between-run variance matching the
fitted value, that fit returned exactly zero in 49\% of replicates. The bootstrap
reported here therefore uses an optimizer-free one-way ANOVA moment estimator,
which admits the boundary and reaches it in about 0.5\% of resamples. The interval
is close to estimator-invariant: the lower bound lies near 0.03 and the upper near
0.74 under this estimator and under a multi-start REML alternative alike. The
point estimates quoted above are REML fits, with which the moment estimator agrees
closely (Arm A bootstrap median 0.54 against a REML point estimate of 0.54).} The
interval is wide, an expected consequence of estimating a between-group variance
from only 13 to 14 groups, and we therefore do not claim a precise value; we
report the interval rather than a bootstrap median, since the median of a
non-negative variance ratio is an unstable summary when a share of resamples sits
near the boundary. What the data support is that a substantial between-run
component is present in both arms: the point estimates place roughly half the
variance between runs, and the bootstrap intervals are consistent with them.
Critically, the modelling consequence that follows does not depend on the
precise figure. Even at the lower end of the plausible range, enough variance lies
between runs that a recipe-only model, whose inputs are constant within a run, is
bounded well below what within-run information could achieve, which is confirmed
directly in Section~\ref{sec:ridge}: the recipe-only model fails to transfer
across runs, its pooled leave-one-batch-out $R^{2}$ only marginally positive.

This single number has a strong modelling consequence. A model whose only inputs
are recipe parameters, which are constant within a run, can by construction explain
only the between-run share of variance, and only to the extent the recipe actually
drives it; it has no access to the within-run half at all. With about half the
variance sitting between runs and the recipe explaining only part of that, a
recipe-only model is inadequate not because it is poorly fit but because the
variance structure forbids it. The decomposition therefore motivates the modelling
choice that follows: a mixed-effects formulation that represents the run explicitly
as a random effect, rather than a single-level model that must pretend the runs are
exchangeable.

\subsection{Mixed-effects forward model and the within-run sign flip}\label{sec:mixed}

We model wafer-level gain $y_{ij}$ for wafer $j$ in run $i$ as
\begin{equation}
y_{ij} = \mathbf{x}_{ij}^{\top}\boldsymbol{\beta} + u_i + v_i\, p_{ij} + \varepsilon_{ij},
\label{eq:mixed}
\end{equation}
where $\mathbf{x}_{ij}$ are the standardized recipe features (emitter dose,
screen-oxide thickness, emitter sheet resistivity and its within-batch spread)
with fixed effects $\boldsymbol{\beta}$, $p_{ij}$ is the within-run position,
$u_i \sim \mathcal{N}(0,\sigma^2_u)$ is the per-run random intercept, $v_i \sim
\mathcal{N}(0,\sigma^2_v)$ the per-run random slope on position, and
$\varepsilon_{ij}\sim\mathcal{N}(0,\sigma^2_\varepsilon)$ the residual. Comparing
nested specifications by likelihood-ratio test selects the model with both a
random intercept \emph{and} a random slope on within-run position (per-run
intercept and slope, unstructured covariance).

The random slope is strongly warranted. A likelihood-ratio test of the
random-slope model against the random-intercept-only model gives $\chi^2 = 60.2$
(Arm A) and 52.1 (Arm B). The two models differ by two parameters: the
random-slope variance, which is constrained to be non-negative and so takes its
null value on the boundary of the parameter space, and the intercept--slope
covariance, which is unrestricted. The usual $\chi^2_2$ reference therefore does
not apply; under these boundary conditions the asymptotic null distribution is a
50:50 mixture, $\tfrac{1}{2}\chi^2_1 + \tfrac{1}{2}\chi^2_2$
\citep[Case~6]{selfliang1987}, which is stochastically smaller than $\chi^2_2$.
The naive $\chi^2_2$ tail probabilities ($\approx 10^{-14}$ for Arm A and
$10^{-12}$ for Arm B) are thus conservative upper bounds on the true $p$-value,
which is smaller still. The random slope is not a marginal addition.

Among the fixed effects, emitter dose and screen-oxide thickness carry the largest
positive standardized coefficients (dose $+46.3$, thickness $+37.8$ in Arm A;
$+43.0$ and $+30.6$ in Arm B, all with $p < 0.005$ in both arms), and the
resistivity spread carries a significant negative coefficient ($-32.7$, $p =
0.037$ in Arm A; $-37.6$, $p = 0.009$ in Arm B). The mean resistivity is not
individually significant in either arm. That dose and thickness dominate is
consistent with the device physics: both act on the emitter, which sets the gain.

The headline finding lives in the position term. For variance explained we use the
marginal and conditional $R^2$ of \citet{nakagawa2013} in the random-slope form of
\citet{johnson2014}; because the response is Gaussian with an identity link, the
distribution-specific variance is zero and the two reduce to
\begin{equation}
R^2_{\text{marginal}} = \frac{\sigma^2_f}{\sigma^2_f + \sigma^2_\alpha + \sigma^2_\varepsilon},
\qquad
R^2_{\text{conditional}} = \frac{\sigma^2_f + \sigma^2_\alpha}{\sigma^2_f + \sigma^2_\alpha + \sigma^2_\varepsilon},
\label{eq:r2}
\end{equation}
where $\sigma^2_f$ is the variance of the fixed-effect predictions and
$\sigma^2_\varepsilon$ the residual variance. Here $\sigma^2_\alpha$ is not a
single random-intercept variance but the mean random-effect variance of the fitted
intercept-and-slope structure, the covariate-averaged term
$\mathrm{Tr}(\mathbf{Z}\boldsymbol{\Sigma}\mathbf{Z}^{\top})/n$ of
\citet{johnson2014}, which is what makes these $R^2$ values well defined once the
model carries a random slope on position. The marginal $R^{2}$ of the selected
model, the variance explained by the fixed effects alone, is 0.468 (Arm A) and
0.466 (Arm B), while the conditional $R^{2}$, fixed plus random effects, is 0.678
and 0.640. The gap between them carries the finding in two numbers: a substantial
share of what the model captures, between a quarter and a third of the explained
variance, is run-specific structure the fixed effects cannot see, which is exactly
the reading \citet{nakagawa2013} give to the marginal--conditional difference,
namely the share of variability residing in the random effects.

\begin{figure}[htbp]
\centering
\includegraphics[width=0.7\textwidth]{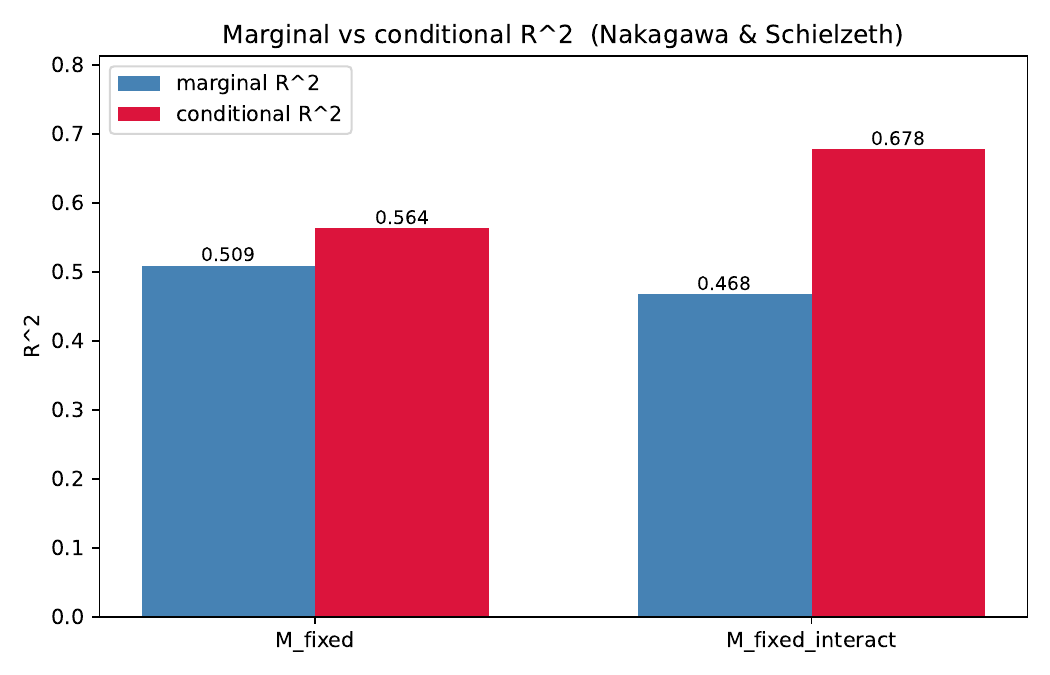}
\caption{Marginal versus conditional $R^2$ for the mixed-effects model, Arm A. The
gap between the two, which is the variance captured by the run-level random
effects rather than the fixed recipe effects, is the variance-hierarchy finding
expressed as a single quantity}
\label{fig:marginal_conditional}
\end{figure}

Adding the random slope moves the two measures in opposite directions: marginal
$R^{2}$ falls relative to the random-intercept model ($0.509 \rightarrow 0.468$ in
Arm A) while conditional $R^{2}$ rises ($0.564 \rightarrow 0.678$ in Arm A; both
fits are shown in Fig.~\ref{fig:marginal_conditional}). The drop in marginal
$R^{2}$ is expected rather than anomalous: \citet{johnson2014} notes that the
fixed-slope estimate, and with it $\sigma^2_f$, can differ between a
random-intercept and a random-slope fit, with the difference small when the design
is balanced but potentially considerable when the random-intercept slope is biased
toward an unrepresentative subset of runs, so marginal $R^{2}$ need not be
preserved when the slope is freed. In our data the direction of that change is the
quantitative counterpart of a position effect that is real within runs but not
shared in sign across them: pooling it into one fixed slope averages the per-run
effects into a single coefficient that is statistically significant yet matches no
individual run. We do not assume that sign inconsistency here; it is established
directly by the per-run analysis below, and the opposing movement of the two
$R^{2}$ values is its summary in one pair of numbers.

\begin{figure}[htbp]
\centering
\includegraphics[width=0.7\textwidth]{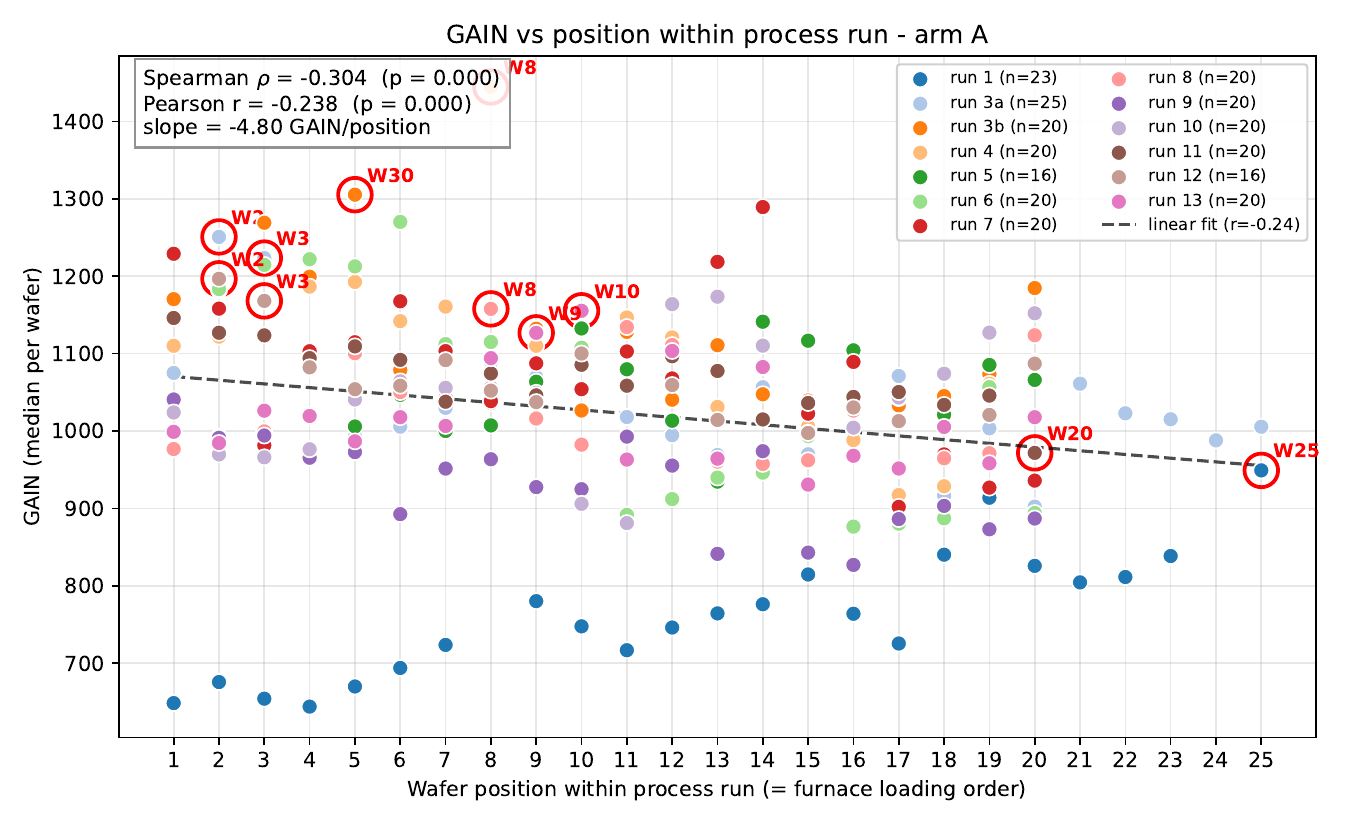}
\caption{Wafer-level gain against within-run position for Arm A. The relationship
motivates the per-run analysis: a position effect is present within runs but does
not share a consistent direction across them}
\label{fig:gain_vs_position}
\end{figure}

\begin{figure}[htbp]
\centering
\begin{minipage}{0.9\textwidth}\centering
\includegraphics[width=\linewidth]{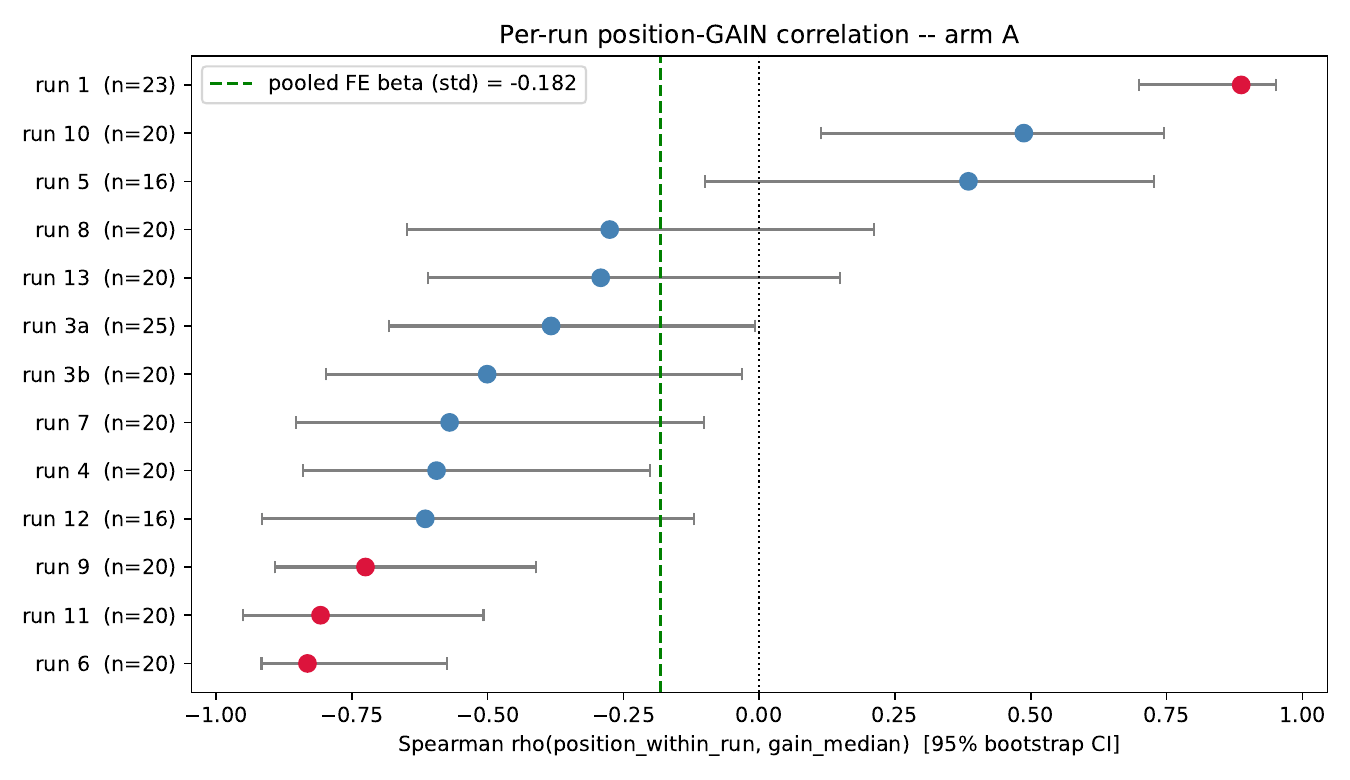}\\[2pt]{\footnotesize (a)}\end{minipage}\\[6pt]
\begin{minipage}{0.46\textwidth}\centering
\includegraphics[width=\linewidth]{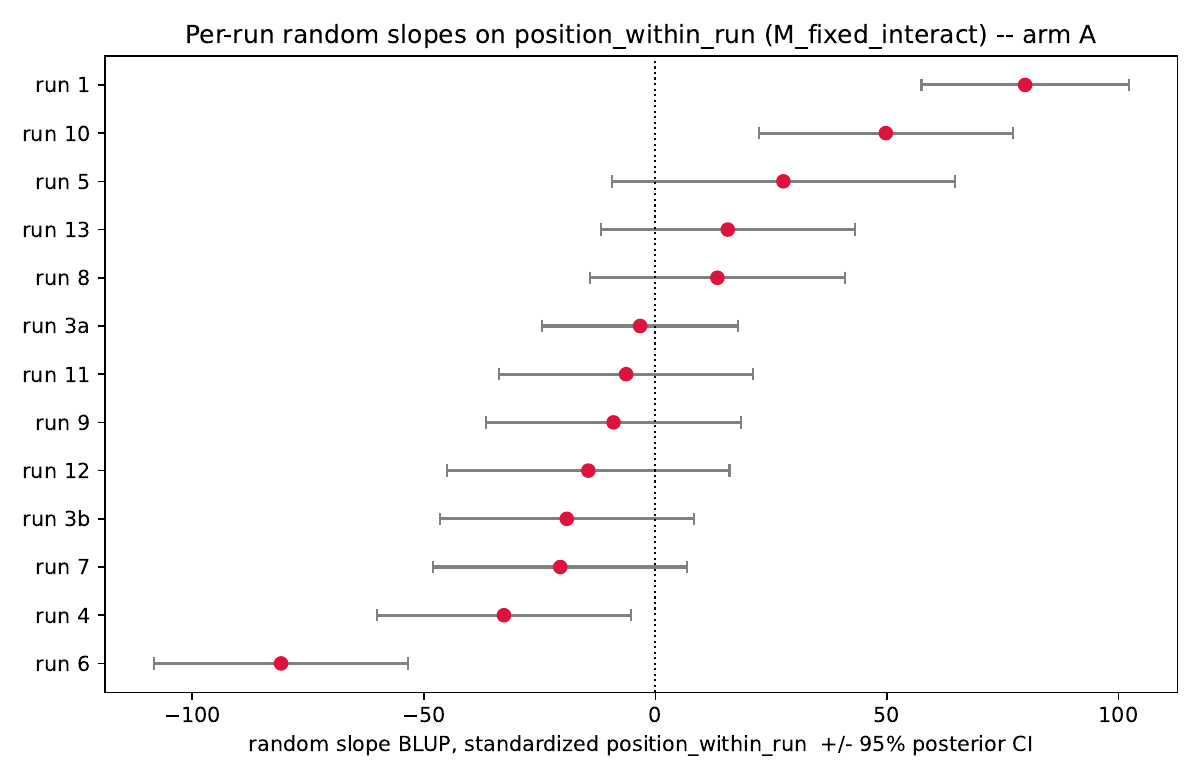}\\[2pt]{\footnotesize (b)}\end{minipage}\hfill
\begin{minipage}{0.46\textwidth}\centering
\includegraphics[width=\linewidth]{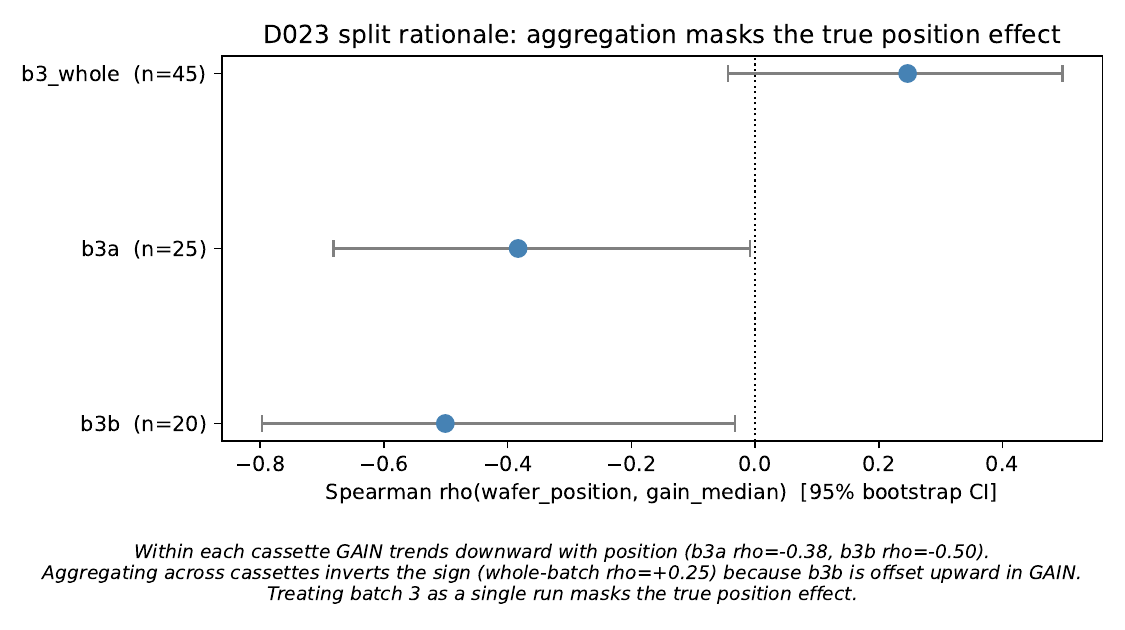}\\[2pt]{\footnotesize (c)}\end{minipage}
\caption{The within-run position effect is real but sign-inconsistent across runs.
\textbf{a} Per-run position--gain rank correlations, with one run strongly
positive and three strongly negative at Bonferroni-corrected significance.
\textbf{b} Random slopes (best linear unbiased predictors) from the mixed-effects
model, which span both signs. \textbf{c} The batch-3 change-point, at which
re-indexing flips the correlation sign. A single pooled coefficient averages these
to a small pooled value that represents no single run}
\label{fig:signflip}
\end{figure}

Figure~\ref{fig:gain_vs_position} plots wafer-level gain against within-run
position and motivates the per-run analysis that follows. We confirm the sign flip
through two genuinely independent statistical views, with two further corroborating
observations, all four collected in Fig.~\ref{fig:signflip}.

First, the pooled fixed-effect position coefficient, the single slope a
recipe-plus-position model would report, is significant and negative ($-3.68$ raw
gain units per position step, $p \approx 10^{-5}$, Arm A; $-3.96$, $p \approx
10^{-6}$, Arm B; about $-23$ to $-26$ in standardized units, the fourth largest of
the five fixed effects). Taken alone it suggests a uniform ``later wafers have
lower gain'' effect, spanning roughly 90 gain units across a 25-wafer run.

Second, the per-run rank correlations tell a different story. Fitting position
against gain within each run separately, four runs reach Bonferroni-corrected
significance in both arms, and they do not agree in sign: run 1 is strongly
positive (Spearman $\rho = +0.89$) while runs 6, 9 and 11 are strongly negative
($\rho = -0.83$, $-0.72$, $-0.81$). A single pooled slope cannot represent one run
rising and three falling; the pooled coefficient is a weighted average that mostly
cancels.

Third, the model's own per-run random slopes (best linear unbiased predictors,
BLUPs) span both signs and recover exactly this: run 1's slope is about $+80$ gain
units while run 6's is about $-80$, in both arms. The per-run slopes from the
mixed-effects model and the independent per-run Spearman correlations agree
closely (Pearson $r \approx 0.88$--0.90 between the two), so the random-slope
structure is not an artefact of the model; it tracks a pattern visible in the raw
within-run data.

Fourth, the batch-3 cassette split (Section~\ref{sec:missing}) is the same
phenomenon caught at a batch boundary: re-indexing position at the boundary flips
the within-segment correlation from a non-significant $+0.25$ to $-0.47$ ($p =
0.001$), with a coherent negative within-half slope ($\rho = -0.50$, $p = 0.025$).
This re-indexing signature does not depend on the boundary being the maximal cut,
so it is unaffected by the boundary having been selected from a scan.

The two independent views (per-run rank correlations and mixed-model random slopes,
agreeing at Pearson $r$ of approximately 0.89), together with the corroborating
pooled coefficient and the change-point re-indexing, are mutually consistent: the
within-run position effect is real and substantial within individual runs, but its
sign is not shared across runs, so any model that forces a single position
coefficient will report a slope that fails to represent any individual run and
understates how much position matters within the runs where the effect is strong.
We do not interpret position as a furnace-slot coordinate; as noted in
Section~\ref{sec:missing} the physical load position is not recorded, so position
is treated as within-run sequence order.

\subsection{Wafer-level ridge cross-check: the marginal value of position}\label{sec:ridge}

To check that these conclusions do not depend on the mixed-effects machinery, we
repeat the analysis with a transparent wafer-level ridge regression, validated by
leave-one-batch-out cross-validation (LOBO-CV). The wafer-level ridge
regression minimizes
\begin{equation}
\hat{\boldsymbol{\beta}} = \arg\min_{\boldsymbol{\beta}} \; \lVert \mathbf{y} - \mathbf{X}\boldsymbol{\beta}\rVert_2^2 + \lambda \lVert\boldsymbol{\beta}\rVert_2^2,
\label{eq:ridge}
\end{equation}
with penalty $\lambda$ selected by leave-one-batch-out cross-validation. This is
the honest test for a process with strong between-run variance: a standard random
split would leak run-level information between train and test and report
optimistic numbers. Table~\ref{tab:ridge} collects the results.

Two results matter. First, the recipe-only picture is confirmed from the other
direction: cross-run generalization is poor. The recipe-only model's pooled
LOBO-CV $R^{2}$ is only marginally positive (0.062 in Arm A, 0.108 in Arm B);
adding within-run position raises the pooled value slightly but drives the
mean-of-folds $R^{2}$ strongly negative ($-1.22$ in Arm A, $-1.06$ in Arm B), the
pooled and per-fold measures diverging because the model captures part of the
between-run offset while having no skill relative to each run's own mean. We report
these values rather than hiding the negative one, because it is the expected
consequence of the variance hierarchy in Section~\ref{sec:variance}, not a
modelling failure. A recipe-only model simply does not transfer across runs.

Second, position carries genuine marginal value. Adding within-run position raises
the pooled leave-one-batch-out $R^{2}$ from 0.062 to 0.108 (Arm A) and from 0.108
to 0.155 (Arm B), a difference of $+0.046$ and $+0.047$. This improvement is
positive in both arms but small relative to its uncertainty: a cluster bootstrap
gives a 95\% interval of $[-0.018, 0.197]$ (Arm A) whose lower bound sits just
below zero, and a one-sided sign-rank test across the held-out-run folds is only
marginally significant ($p = 0.06$ in Arm A, $p = 0.12$ in Arm B). We therefore
characterize position as adding predictive information on average rather than as a
decisively positive effect, consistent with the finding of
Section~\ref{sec:mixed} that its direction varies across runs: a feature whose
sign is run-specific improves prediction unevenly and need not register as
uniformly significant under leave-one-run-out evaluation. That the same feature is
both small relative to its per-run spread as a pooled \emph{coefficient}
(Section~\ref{sec:mixed}) and yet positive on average as a marginal
\emph{predictor} is not a contradiction; it reflects that position matters within
runs even though its direction varies, so it improves prediction without resolving
into a clean global sign.

The wafer-level ridge model does not store an interpretable coefficient vector in
the released results; the mixed-effects fixed-effects table of
Section~\ref{sec:mixed} serves as the interpretable coefficient summary for the
forward model, and the ridge model is used here specifically as a
model-class-independent cross-check on the LOBO-CV generalization and the marginal
value of position.

\begin{table}[htbp]
\centering
\caption{Wafer-level ridge regression: pooled leave-one-batch-out (LOBO-CV)
$R^{2}$, the incremental $\Delta R^{2}$ from adding within-run position, and the
random-forest comparison. The mean-of-folds $R^{2}$ is reported as obtained and is
strongly negative, which is the expected consequence of the between-run variance
hierarchy rather than a fitting failure. Ridge forward-model root-mean-square
error on held-out runs is approximately 115 gain units in Arm A and 111 in Arm B.
No in-sample training $R^{2}$ is stored, so models are compared on cross-validated
performance only}
\label{tab:ridge}
\setlength{\tabcolsep}{4pt}
\begin{tabular}{lrrrr}
\toprule
& \multicolumn{2}{c}{Arm A} & \multicolumn{2}{c}{Arm B} \\
\cmidrule(lr){2-3}\cmidrule(lr){4-5}
Model & pooled $R^2$ & mean-of-folds $R^2$ & pooled $R^2$ & mean-of-folds $R^2$ \\
\midrule
Ridge, recipe only (no position) & 0.062 & --- & 0.108 & --- \\
Ridge, with within-run position & 0.108 & $-1.22$ & 0.155 & $-1.06$ \\
$\Delta R^2$ (position) & $+0.046$ & & $+0.047$ & \\
\midrule
Random forest (with position) & $-0.28$ & $-2.38$ & $-0.22$ & $-2.15$ \\
\bottomrule
\end{tabular}
\end{table}

\subsection{Forward prediction: what the model delivers and which features drive it}\label{sec:forward}

The first part of this section established the variance structure and the
within-run position finding. This second part turns the same models toward
prediction: what an engineer can forecast before fabrication, what recipe the
models recommend for a target gain, and how much each prediction can be trusted.

The deployable forward model is the mixed-effects model of
Section~\ref{sec:mixed}, with the wafer-level ridge regression of
Section~\ref{sec:ridge} as a transparent cross-check, whose leave-one-batch-out
performance is reported there. The value of the forward model is not high point
accuracy, which the variance hierarchy of Section~\ref{sec:variance} caps, but an
honest forecast that, combined with the uncertainty quantification of
Section~\ref{sec:gp}, tells an engineer both a predicted gain and how far to trust
it.

\begin{figure}[htbp]
\centering
\includegraphics[width=0.8\textwidth]{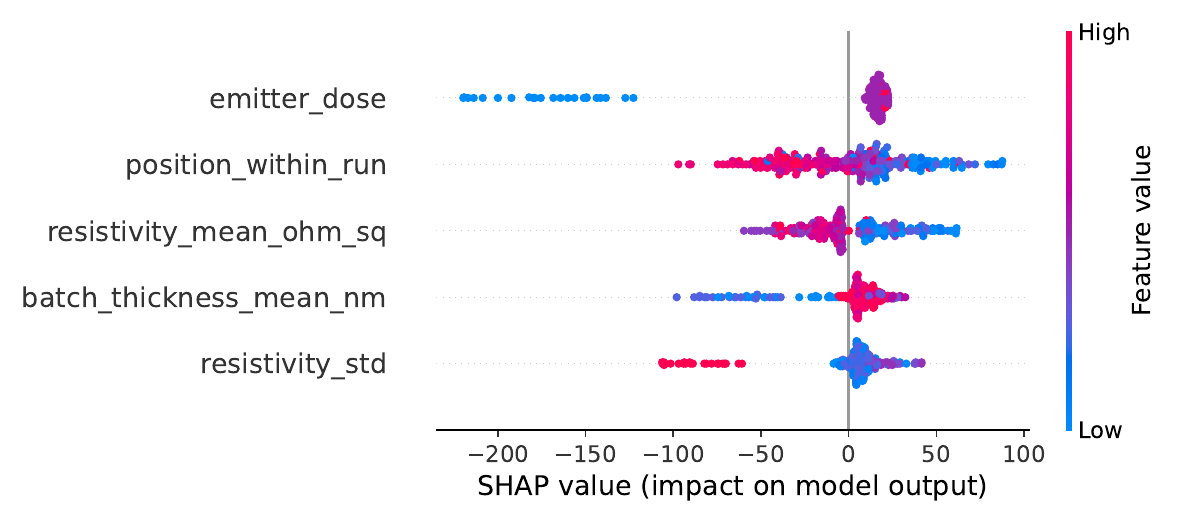}
\caption{SHAP summary for the wafer-level random-forest fit, Arm A, shown as an
in-sample diagnostic of feature associations rather than an attribution of the
deployed model. Emitter dose and within-run position carry the largest
attributions}
\label{fig:shap}
\end{figure}

Which features drive that prediction is answered first by the model we actually
deploy: the mixed-effects fixed effects of Section~\ref{sec:mixed}. Two
independent importance analyses corroborate the ordering reported there. Both are computed
on a random-forest\footnote{A random forest is an ensemble machine-learning model
that, for regression, averages the predictions of many decision trees. Each tree is
grown on a bootstrap resample of the data, and at every split a fresh random subset
of the features is considered as split candidates; this per-node feature sampling
decorrelates the trees, and averaging over them yields a flexible non-linear
regressor.} fit \citep{breiman2001} and explained in-sample, so we read them as
descriptive of the associations present in the data rather than as attributions of
the deployed model, a distinction that matters because the random forest itself
does not generalize across runs (Section~\ref{sec:selection}). With that caveat,
they agree with the mixed-effects model and with each other: permutation importance
ranks emitter dose and within-run position as the two strongest features in both
arms (mean $\Delta R^{2} \approx 0.47$ and 0.46 in Arm A; in Arm B the order
reverses, position 0.45 and dose 0.33), followed by the two resistivity features
and thickness; and SHAP\footnote{SHAP (SHapley Additive exPlanations) is a
feature-attribution framework that uses the game-theoretic Shapley value to assign
each feature a contribution to a given prediction, quantifying how much each
feature moves the prediction above or below its baseline. It can be estimated
model-agnostically (Kernel SHAP) or with faster model-specific approximations such
as Linear, Max and Deep SHAP.} attributions \citep{lundberg2017}, summarized in
Fig.~\ref{fig:shap}, return the same leading set: emitter dose, within-run position
and mean resistivity. That a linear mixed-effects model, a permutation-importance
analysis on a random forest, and a game-theoretic attribution method independently
elevate the same few features is the useful result: the forward model's dependence
on dose, position and the emitter-defining parameters is not an artefact of one
method. We note that within-run position ranks highly here as an \emph{association}
even though, as Section~\ref{sec:mixed} showed, its pooled coefficient is
significant but much smaller than the per-run slopes of about $\pm 80$ because its
sign varies across runs; the importance methods detect that position carries
information without resolving its direction.

\subsection{Inverse recipe search: recommending a recipe for a target gain}\label{sec:inverse}

The forward model is invertible: given a target gain, we search the recipe space
for the settings predicted to achieve it. We do this directly, by evaluating the
mixed-effects forward model across a grid of candidate recipes (3{,}750 points
spanning the three normalized emitter-dose set-points, fifty resistivity values
and the twenty-five within-run positions, with thickness and resistivity spread
held at their central values) and retaining the recipes whose marginal prediction
band intersects the target band. The prediction band is the fixed-effects
prediction plus or minus one standard error, a roughly 68\% band, not a 95\%
interval. The grid is evaluated as a vectorized batch prediction, so a full sweep
across target gains is immediate; it does not require a separately trained
surrogate model. We report the procedure and its admissible regions; the released
results do not record an execution-time benchmark, so we make no quantitative speed
claim.

\begin{figure}[htbp]
\centering
\includegraphics[width=0.8\textwidth]{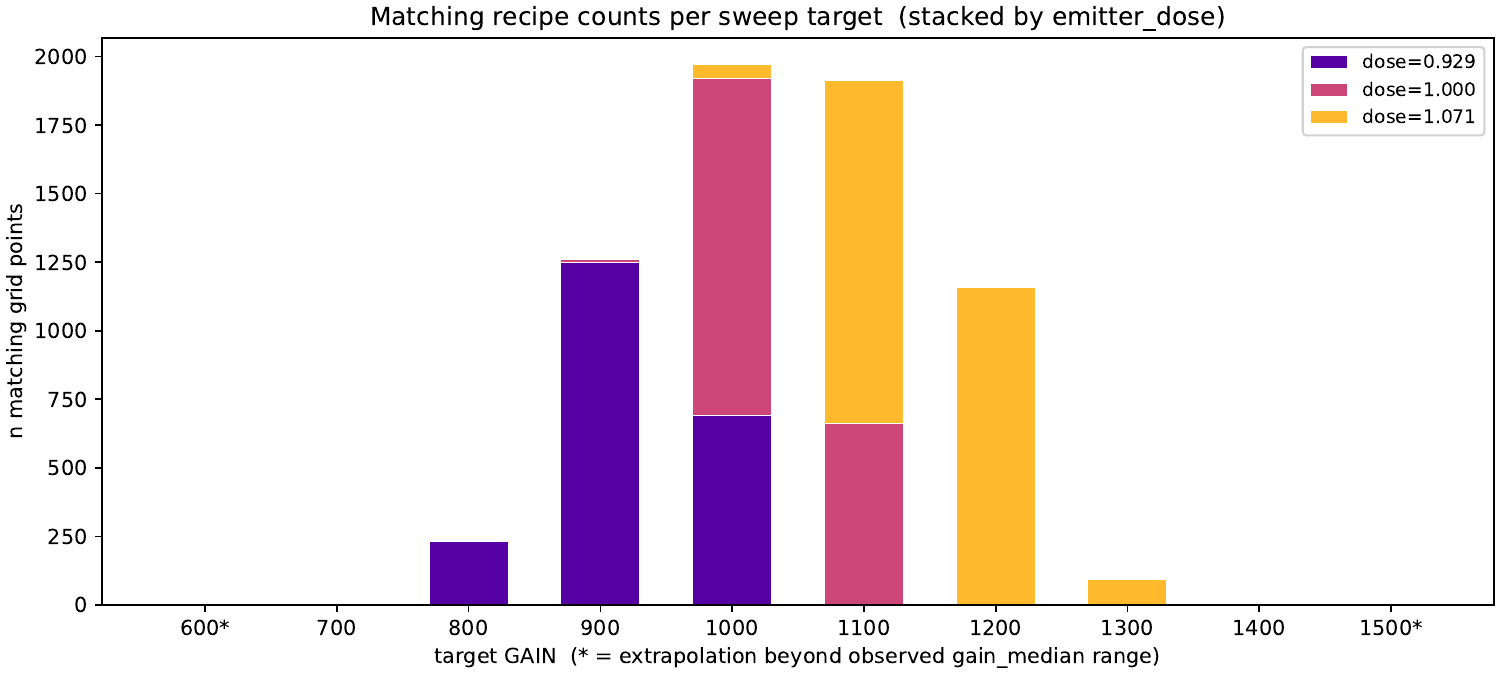}
\caption{Inverse search: number of admissible recipes as a function of target gain
for Arm A with a $\pm 5\%$ band. The count is unimodal, zero at the extremes and
peaking near a target of 1000, giving a feasibility map of reachable gains}
\label{fig:inverse}
\end{figure}

Run across the full target range with a $\pm 5\%$ band, the search recovers a
coherent and physically sensible picture of which gains are achievable, shown in
Fig.~\ref{fig:inverse}. The number of admissible recipes is unimodal in the target:
zero at 600 and 700, rising sharply to a peak near a target of 1000 (about
1{,}970--1{,}985 admissible recipes), and falling back to zero by 1400--1500. In
other words, the model reports that mid-range gains are broadly achievable across
many recipes while the extremes of the device-acceptance window are reachable by
few recipes or none, exactly the kind of feasibility map a process engineer needs
before committing a run.

Position here is treated as within-run sequence order, as in
Section~\ref{sec:mixed}; this is the working interpretation of the position index
and is not independently confirmed in the process records
(Section~\ref{sec:missing}). Because the inverse search uses
the population-average (fixed-effect) position coefficient rather than any single
run's random slope, a reported position reflects the average across runs and is not
a guarantee for one future run: as Section~\ref{sec:mixed} showed, the realized
position effect varies in sign from run to run. We therefore report position as a
recommendable setting only at the population level, while making explicit that its
run-specific realization cannot be predicted in advance.

Within an achievable target, the search exposes a three-tier identifiability
structure in the emitter dose. At a representative target of 1100 ($\pm 5\%$), the
low dose set-point (normalized 0.929) admits no recipe in either arm; the mid
set-point (1.000) admits an intermediate fraction (662 of 1{,}250 grid recipes in
Arm A, 682 in Arm B); and the high set-point (1.071) admits all 1{,}250. The three
tiers have distinct operational meaning. The low dose is \emph{eliminative}: it
cannot reach the target, so observing this target rules it out. The mid dose is
\emph{identifying}: only some combinations of the other parameters reach the
target, so the target meaningfully constrains the recipe. The high dose is
\emph{saturated}: it reaches the target across the whole sub-grid, so the target
tells you little about the remaining parameters. This structure is itself useful
guidance, since it tells an engineer that at the high dose the target is easy but
under-determined, while at the mid dose hitting the target pins down the recipe
more tightly. That the emitter implant dominates the recipe in this way is
consistent with independent device-physics evidence: in polysilicon-emitter bipolar
transistors, the emitter--base interface oxide, whose thickness varies with furnace
boat position (top versus bottom of the tube), is reported to be a controlling
factor on current gain \citep{oxide_ieee}, an emitter-side sensitivity our model
recovers from data alone.

\subsection{Gaussian-process uncertainty and engineer-facing trust tiers}\label{sec:gp}

A forward prediction is only actionable if it carries an honest uncertainty. We fit
a Gaussian-process regressor (Mat\'ern kernel with $\nu = 2.5$ plus a white-noise
term) over the recipe features and use its predictive standard deviation as the
basis for engineer-facing trust tiers. The GP excludes the resistivity spread as a
feature because its fitted length scale ran to the boundary, so the GP treats it as
effectively constant; the GP therefore operates on emitter dose, mean resistivity
and thickness.

\begin{figure}[htbp]
\centering
\includegraphics[width=0.75\textwidth]{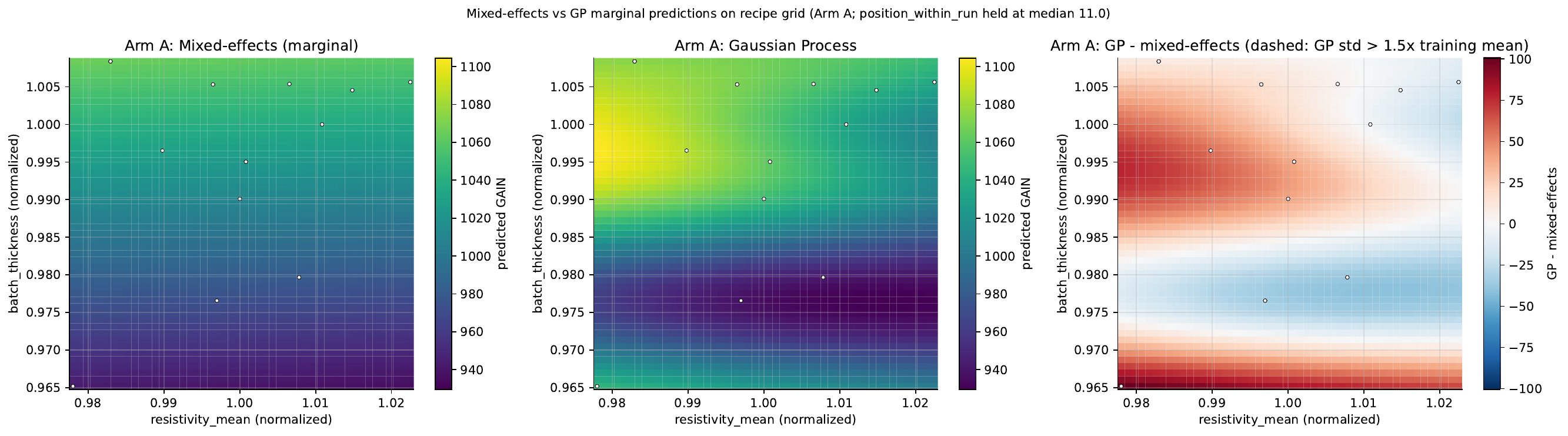}
\caption{Gaussian-process versus mixed-effects predictions for Arm A. The two model
families agree closely on interpolation within observed runs, supporting the
interpretation that agreement under cross-validation and divergence under
leave-one-batch-out validation reflect the between-run variance rather than a
model defect}
\label{fig:gp_vs_me}
\end{figure}

The GP's behaviour confirms and sharpens the variance-hierarchy story. Under
five-fold cross-validation its $R^{2}$ is 0.501 (Arm A) and 0.447 (Arm B), close to
the mixed-effects model's marginal $R^{2}$ (0.468 and 0.466) and consistent with
the ridge cross-check, so on interpolation within the observed runs the three model
families agree on how much variance is explainable; Fig.~\ref{fig:gp_vs_me}
compares the Gaussian-process and mixed-effects predictions directly. Under
leave-one-batch-out validation, however, the GP's $R^{2}$ falls to strongly
negative (mean-of-folds $-1.79$ and $-1.69$). On the same mean-of-folds metric the
wafer ridge is also strongly negative ($-1.22$ and $-1.06$), so both model families
lose predictive skill when asked to generalize to an entirely unseen run; the GP is
the worse of the two by about 0.6 in $R^{2}$. We compare the two on the mean-of-folds
metric because a pooled $R^{2}$ is recorded for the ridge but not for the GP; the
ridge's pooled value is positive ($+0.108$ and $+0.155$), and the contrast between
that pooled figure and the strongly negative per-fold figures reflects the
difference between the two aggregation schemes rather than a difference between the
models. This is not a controlled model comparison in any case, since the GP operates
on three recipe features and the ridge on five. This pattern, agreement under cross-validation and divergence under
leave-one-batch-out, is not a defect: it is the uncertainty model correctly
reporting that predicting an entirely unseen run is far harder than interpolating
within seen runs, the signature of the between-run variance established in
Section~\ref{sec:variance}.

\begin{figure}[htbp]
\centering
\includegraphics[width=0.8\textwidth]{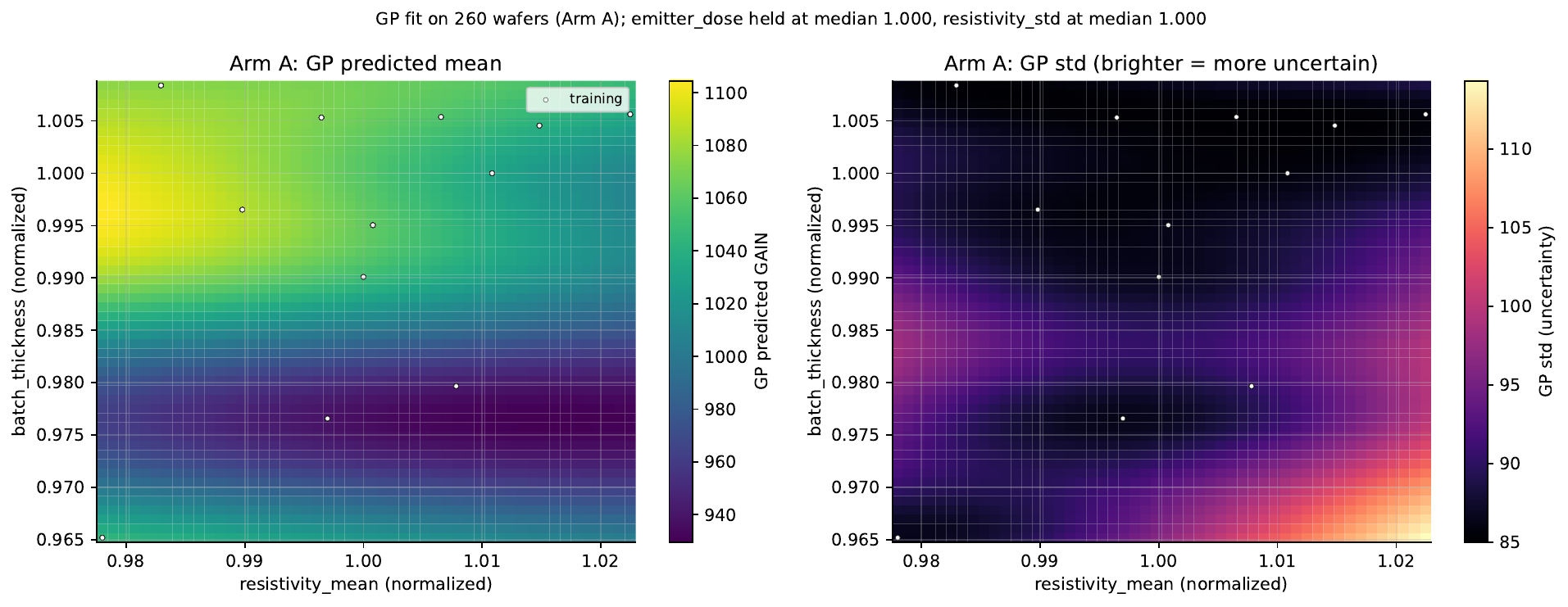}
\caption{Gaussian-process predictive-uncertainty map over the recipe space for Arm
A, with trust tiers defined against the training-region spread: HIGH at or below
$1.2\times$, MEDIUM at or below $1.5\times$, and LOW above $1.5\times$. LOW-trust
regions flag recipes far from anything fabricated}
\label{fig:gp}
\end{figure}

That divergence is what makes the GP useful for deployment rather than just for
scoring. We translate its predictive standard deviation into three tiers against
the typical in-sample spread: a prediction is HIGH trust when the GP standard
deviation is within $1.2\times$ the training-region spread (a data-dense region),
MEDIUM up to $1.5\times$ (the edge of the training data), and LOW beyond that (an
extrapolation zone). The reference spread is about 86--88 gain units. Surfaced
alongside a predicted gain, and mapped over the recipe space in
Fig.~\ref{fig:gp}, these tiers give an engineer a direct, legible signal: a
HIGH-trust prediction sits where the model has seen comparable recipes, while a
LOW-trust prediction flags a recipe far from anything fabricated, where the
forecast should be treated as a rough indication and verified.

\subsection{Model selection and honest limits}\label{sec:selection}

Our model choices follow from the small-sample, hierarchical setting rather than
from a search for the highest in-sample fit. At the wafer level we use ridge
regression and a random forest, evaluated by leave-one-batch-out cross-validation.
The random forest generalizes worse than ridge on every cross-validated metric: its
pooled LOBO-CV $R^{2}$ is negative ($-0.28$ in Arm A, $-0.22$ in Arm B) where ridge
is positive ($+0.11$, $+0.16$), and its mean-of-folds $R^{2}$ is more negative than
ridge's ($-2.38$ versus $-1.22$ in Arm A; $-2.15$ versus $-1.06$ in Arm B). We
therefore prefer ridge as the wafer-level cross-check: a flexible, high-capacity
learner has no advantage when the binding constraint is between-run variance rather
than functional complexity, and its extra capacity simply fits run-specific noise
that does not transfer. The released results do not store an in-sample training
$R^{2}$ for these models, so we compare them on cross-validated performance only,
not on a train-versus-test gap; the conclusion rests on the random forest's worse
cross-validated generalization.

The honest limits follow directly. The forward model's cross-run $R^{2}$ is low by
construction, not by under-fitting (Section~\ref{sec:variance}). The uncertainty
tiers promise a relative signal, denser versus sparser regions of recipe space, not
a calibrated coverage guarantee; a HIGH-trust label means that comparable recipes
were seen, not that the true gain falls in a stated interval with a stated
probability. The inverse search returns recipes the \emph{model} predicts will hit
a target, inheriting all of the forward model's uncertainty; it is a screening aid,
not a guarantee of a fabricated outcome. And every result rests on thirteen to
fourteen runs of one device class; the structure we report (variance hierarchy,
sign-flipping position effect, dose identifiability tiers) is what the data
support, and its generalization to other devices is a hypothesis for future work,
not a claim made here. And, considering the unclear impact and meaning of the
position of wafers in the furnaces, the systematic recording of how wafers are
loaded in each batch will be recommended.

\section{Engineer-facing assistant}\label{sec:assistant}

The forward, inverse and uncertainty models of Section~\ref{sec:models} are useful
to a process engineer only if they can be queried without the statistical tooling
that produced them. We therefore provide a natural-language assistant that lets an
engineer ask, in plain language, either a forward question (what gain would this
recipe give?) or an inverse one (what recipe reaches this gain?) and receive an
answer drawn from the validated models. The assistant is a deployment layer over
Section~\ref{sec:models}, not a new model: it adds accessibility, not predictive
power.

\subsection{Architecture}\label{sec:assistant-arch}

The assistant couples a local, open-weights large language model (Llama~3.1, served
through Ollama) to the validated models through a small set of typed tools. Two
tools are exposed: a forward tool that takes a recipe and returns the mixed-effects
prediction of Section~\ref{sec:mixed} together with the Gaussian-process standard
error and trust tier of Section~\ref{sec:gp}, and an inverse tool that takes a
target gain and returns the admissible recipes from the search of
Section~\ref{sec:inverse}, grouped by the dose-identifiability tiers reported there.
The language model's role is deliberately narrow: it interprets the question and
decides which tool to call and with what arguments, the tool then executes the
validated model in ordinary code, and the language model phrases the returned
values back to the engineer. The language model itself performs no arithmetic and
fits nothing. A query may be directed to either analysis arm
(Section~\ref{sec:dataset}), so the imputation sensitivity carried through the
paper remains visible at the point of use. The whole system runs on a single local
machine, with no query or recipe transmitted to an external service, which is a
precondition for use on proprietary fabrication data.

\subsection{Provenance and the absence of fabricated numbers}\label{sec:assistant-prov}

The central design property follows from this separation. Because every gain value,
every recipe and every trust tier originates from a model call rather than from the
language model's own parameters, the assistant cannot fabricate a numerical result:
asked a quantitative question it has no tool for, the language model is constrained
to request the missing inputs rather than to invent an answer. Each response
carries a provenance record showing which tool was called, the arguments it
received and the values it returned (Fig.~\ref{fig:assistant}b), so the engineer can
confirm that the numbers were produced by the validated models. This makes the
honesty the rest of the paper argues for a structural property of the interface
rather than a matter of trust in the language model: the natural-language layer may
phrase an answer imperfectly, but it cannot manufacture a gain or a recipe. Because
the numbers come from the deterministic models, the substantive content of an
answer is reproducible even though the language model's wording is not.

\subsection{Demonstration and scope}\label{sec:assistant-demo}

\begin{figure}[htbp]
\centering
\begin{minipage}{0.92\textwidth}\centering
\includegraphics[width=\linewidth]{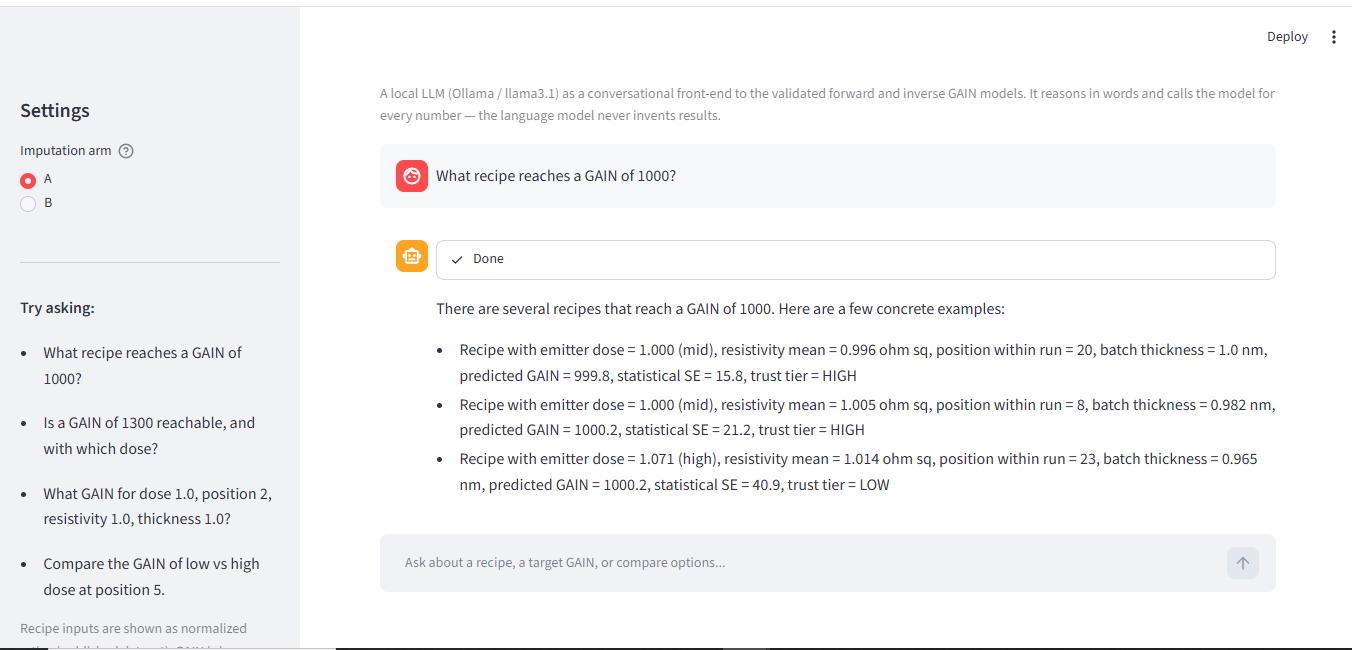}\\[2pt]{\footnotesize (a)}\end{minipage}\\[6pt]
\begin{minipage}{0.92\textwidth}\centering
\includegraphics[width=\linewidth]{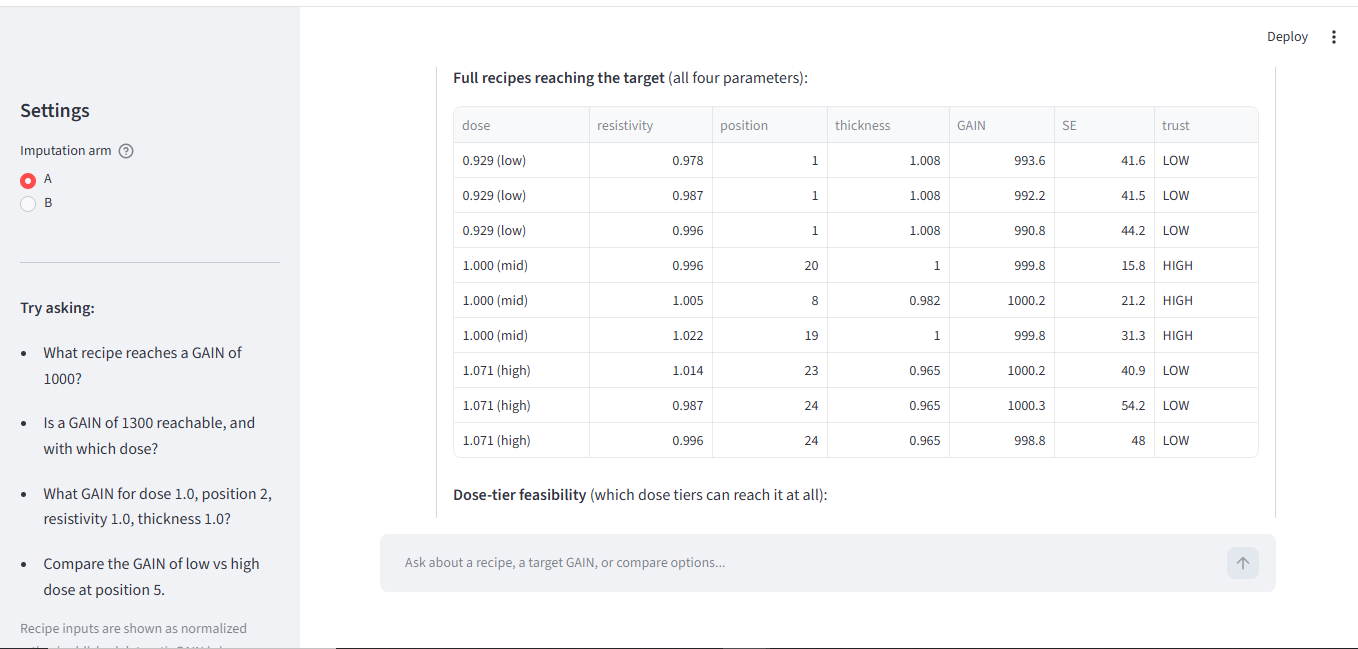}\\[2pt]{\footnotesize (b)}\end{minipage}
\caption{The engineer-facing natural-language assistant. \textbf{a} An inverse
query, in which the engineer asks in plain language for a recipe reaching a target
gain of 1000; a local language model (Llama~3.1, served through Ollama) interprets
the request, invokes the inverse search of Section~\ref{sec:inverse}, and reports
complete recipes with predicted gain, standard error and trust tier. \textbf{b} The
accompanying provenance panel, in which every value is produced by the validated
models rather than the language model, listing the tool called, the arguments it
received and the recipes returned across the feasible dose tiers. Recipe inputs are
shown as normalized ratios following the released-dataset convention, and gain is in
measured units}
\label{fig:assistant}
\end{figure}

Figure~\ref{fig:assistant} shows the assistant answering an inverse query. For a
target gain of 1000 it returns complete recipes spanning the feasible dose tiers,
each annotated with predicted gain, standard error and trust tier, reproducing in
interactive form the same dose-identifiability structure and trust-tier semantics
established in Sections~\ref{sec:inverse} and~\ref{sec:gp}. The assistant is a
demonstrated proof of concept: its purpose is to show that the validated models can
be made usable by a non-specialist behind a provenance guarantee, not to claim a
contribution in language modelling. Its conversational robustness has not been
evaluated in a controlled study, and it inherits every limitation of the models it
wraps (Section~\ref{sec:selection}).

\section{Discussion}\label{sec:discussion}

\subsection{Manufacturing implications}\label{sec:manuf}

The practical motivation for this work is the cost asymmetry involved in
customizing recipes and process flows: a phototransistor run commits months of
cleanroom time and irreversible material before any device can be measured, so any
tool that screens recipes \emph{before} a run has leverage out of proportion to its
accuracy. The forward and inverse models developed here are that tool. Given a
candidate recipe they return a predicted gain with a relative trust tier, and given
a target gain they return the set of recipes the model judges capable of reaching
it, both in the time of a grid evaluation rather than a fabrication cycle. We are
deliberate about what this buys. The cross-run predictive accuracy is modest by
construction, a pooled leave-one-batch-out $R^{2}$ of roughly 0.11--0.16, and, as
Section~\ref{sec:variance} establishes, this is the direct consequence of the
variance hierarchy rather than a shortfall to be engineered away. The contribution
is therefore an \emph{honest screening} capability: the models flag, through the
inverse search's admissible-region map and the Gaussian-process trust tiers, which
targets are broadly reachable and which predictions sit far from any fabricated
recipe. Used this way, as a funnel that removes clearly infeasible recipes and
ranks the rest, even a bounded model converts months of blind iteration into a few
informed runs.

\subsection{What the finding means, and what generalizes}\label{sec:generalize}

The scientific finding is that gain in this process is governed as much by
between-run structure as by the recipe, and that the within-run position effect is
real but sign-inconsistent across runs. The variance hierarchy, a between-run
component estimated near half the total though with wide uncertainty at this sample
size, is the load-bearing result: it explains why recipe-only models fail, why a
mixed-effects formulation is mandatory, and why the honest predictive ceiling is
where it is. The position sign flip, supported by two independent statistical views
and corroborated again at the batch-3 cassette boundary, shows that a single pooled
slope can actively mislead, averaging a real effect to a small, unrepresentative
value, and that the structure only becomes visible when the run is modelled
explicitly. What transfers is the \emph{methodology}: the discipline of decomposing
variance before selecting a model class, of treating a hierarchical dataset with
per-level quality dimensions and an explicit cross-level linkage score, and of
replacing silent data omissions with exclusion-with-reason. Any fabrication process
with batch, wafer and die strata and small batch-level sample size faces the same
structural problems this approach was built for; the multi-level data-quality
framework and the variance-decomposition-first modelling sequence are the parts we
expect to carry to other industrial hierarchies, even where the physics and the
numbers differ entirely.

\section{Conclusion and future work}\label{sec:conclusion}

We have studied the prediction and inversion of phototransistor gain from
fabrication process parameters on a small production corpus, and found that the
dominant structure in the data is between-run rather than recipe-driven: roughly
half of gain variance lies between process runs, and the within-run position effect
changes sign across runs so that a pooled model averages it away. These findings
explain why a recipe-only predictor is bounded, and they are recovered only by
modelling the run explicitly through hierarchical random effects. On this basis we
provide a forward gain predictor with a relative, Gaussian-process-based
uncertainty signal and an inverse recipe search, as a screening aid whose accuracy
is the expected consequence of the variance structure. Underpinning the modelling,
we develop a multi-level data-quality assessment tailored to the nested physical
entities of fabrication, with an explicit cross-level linkage score, and we release
a normalized dataset and runnable pipeline for full reproducibility. The
transferable contribution is methodological: decompose variance before choosing a
model, score data quality across physical levels with explicit cross-level linkage,
and surface scarcity and exclusions rather than hiding them.

New datasets on bipolar phototransistors are expected in the next two years,
supporting further validation of the proposed methodology. Furthermore, given the
multi-level structure of the device, another exploitation of this work will be in
considering different devices and the characterization parameters on which they
should be optimized.

Beyond the present corpus of data, the validated models are exposed through a local
natural-language assistant that makes them usable without machine-learning
expertise while guaranteeing that every number it reports is model-derived, and we
expect the multi-level, cross-level-linked data-quality approach to be the
component most readily carried to other industrial hierarchies with the same nested
physical structure and small batch-level samples.

\backmatter

\bmhead{Supplementary information}

Online Resource~1 (ESM\_1.pdf) contains two parts. The first plots the per-wafer
die count against batch with the L3 plausibility band of Eq.~\eqref{eq:band}
overlaid, showing the three count-anomaly tiers discussed in
Section~\ref{sec:mldq-applied} as distinct clusters above the dense in-band
population. The second reports the headline results of Section~\ref{sec:models}
recomputed with physical batch~3 treated as a single process run rather than
split, the sensitivity analysis referred to in Section~\ref{sec:missing}.

\bmhead{Acknowledgements}

The phototransistor process data and measurements used in this paper refer to
devices fabricated in the Fondazione Bruno Kessler cleanrooms as part of a service
provided to OPTOI SRL. The authors thank Alfredo Maglione, OPTOI CEO, for
authorizing the use of these data.

\bmhead{Use of generative AI and AI-assisted technologies}

During the preparation of this work the authors used Claude (Anthropic), ChatGPT
(OpenAI) and Grok (xAI) for English language editing and revision of the
manuscript, and to assist with analysis code development. The tools were not used
to generate scientific content, to produce or analyse data, or to formulate any
result, finding or interpretation reported here. All output was reviewed and
verified by the authors, who take full responsibility for the content of the
publication.

\section*{Statements and Declarations}

\subsection*{Funding}
This work was carried out using internal institutional resources. No external
funding, public or private, was received for this study, and no project or grant
supported it. The fabrication service provided to OPTOI SRL, noted below, was a
commercial service and did not constitute research funding or support for this
study.

\subsection*{Competing interests}
The devices analysed in this work were fabricated at Fondazione Bruno Kessler as
part of a service provided to OPTOI SRL, which authorized the use of the data for
this publication. Paolo Conci is the founder of Microfab Solutions; he contributed
to this work as an unpaid research collaborator, and no payment or other
consideration was made or requested between Microfab Solutions and Fondazione
Bruno Kessler in connection with this study. The authors declare no other
competing interests.

\subsection*{Ethics approval}
Not applicable. This study involved no human participants and no animals.

\subsection*{Consent to participate and consent for publication}
Not applicable.

\subsection*{Data availability}
The normalized wafer-level dataset supporting the findings of this study is openly
available in Zenodo under a CC-BY-4.0 licence at
\url{https://doi.org/10.5281/zenodo.21379925} \citep{amirabgir2026data}. The raw
die-level records and the per-arm normalization constants are not publicly
available because they are proprietary process data held under Fondazione Bruno
Kessler's data-sharing policy; they are not required to reproduce any result
reported here, since all modelling operates on wafer-level summaries and the
normalization leaves every model's predictions unchanged (the penalized and
tree-based models standardize features within the training pipeline).

\subsection*{Code availability}
The analysis code and trained models are openly available on GitHub at
\url{https://github.com/mahshid-amirabgir/phototransistor-virtual-metrology}
(release v1.1.0), together with the normalized dataset and a dual-mode pipeline
that reproduces every figure and number reported here.

\subsection*{Materials availability}
Not applicable.

\subsection*{Author contributions}
Conceptualization: Mahshid Amirabgir, Lorenza Ferrario, Paolo Conci,
Giancarlo Orengo;
Methodology: Mahshid Amirabgir, Lorenza Ferrario, Giancarlo Orengo;
Software: Mahshid Amirabgir, Mahdieh Amirabgir;
Validation: Mahshid Amirabgir, Lorenza Ferrario, Mahdieh Amirabgir, Paolo Conci,
Giancarlo Orengo;
Formal analysis: Mahshid Amirabgir, Mahdieh Amirabgir;
Investigation: Lorenza Ferrario;
Resources: Mahshid Amirabgir, Lorenza Ferrario;
Data curation: Mahshid Amirabgir, Lorenza Ferrario;
Visualization: Mahshid Amirabgir, Mahdieh Amirabgir;
Writing -- original draft: Mahshid Amirabgir, Lorenza Ferrario,
Mahdieh Amirabgir;
Writing -- review and editing: Mahshid Amirabgir, Lorenza Ferrario,
Mahdieh Amirabgir, Paolo Conci, Giancarlo Orengo;
Supervision: Lorenza Ferrario, Paolo Conci, Giancarlo Orengo;
Project administration: Lorenza Ferrario, Giancarlo Orengo.
All authors read and approved the final manuscript.

\subsection*{ORCID iDs}
Mahshid Amirabgir 0009-0006-4849-2624;
Lorenza Ferrario 0000-0003-2175-4306;
Paolo Conci 0000-0003-4011-7695;
Mahdieh Amirabgir 0009-0005-0615-1058;
Giancarlo Orengo 0000-0003-1044-2999.

\bibliography{refs}

\end{document}